\pdfoutput=1

\documentclass[11pt]{article}

\usepackage{acl}

\usepackage{times}
\usepackage{latexsym}
\usepackage{booktabs}

\usepackage{booktabs, multirow} 
\usepackage{soul}
\usepackage{xcolor,colortbl} 
\usepackage{changepage,threeparttable} 

\usepackage{array}

\usepackage{rotating}
\usepackage{makecell}

 \author{Longju Bai$^1$\thanks{Longju Bai, Angana Borah, and Oana Ignat contributed equally to the manuscript.} \hspace{5pt} 
 Angana Borah$^1{^*}$\hspace{5pt}  
 Oana Ignat$^2{^*}$ \hspace{5pt}
Rada Mihalcea$^1$ \\
$^1$University of Michigan - Ann Arbor, USA  \\
$^2$Santa Clara University - Santa Clara, USA  \\
\textit{\{longju, anganab, mihalcea\}@umich.edu} \hspace{5pt} \textit{oignat@scu.edu} \\  }

\makeatletter
\newcommand\footnoteref[1]{\protected@xdef\@thefnmark{\ref{#1}}\@footnotemark}
\makeatother

\usepackage[utf8]{inputenc}

\usepackage{microtype}

\usepackage{inconsolata}
\usepackage{graphicx}
\usepackage{amsmath}
\usepackage{cleveref}

\graphicspath{{img/}}

\title{\includegraphics[width=0.035\textwidth]{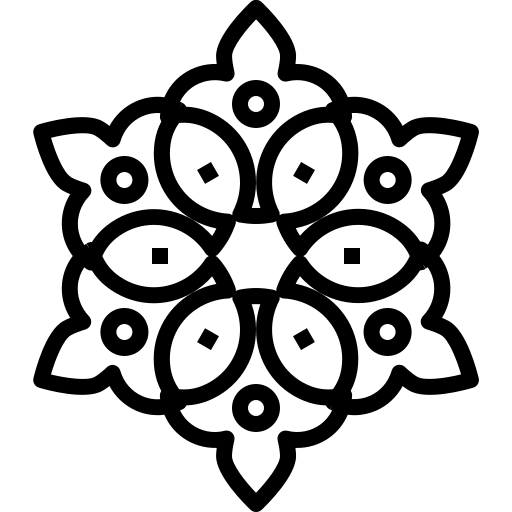} The Power of Many: Multi-Agent Multimodal Models\\ for Cultural Image Captioning}

\begin{document}
\maketitle

\begin{abstract}
Large Multimodal Models (LMMs) exhibit impressive performance across various multimodal tasks. However, their effectiveness in cross-cultural contexts remains limited due to the predominantly Western-centric nature of most data and models. Conversely, multi-agent models have shown significant capability in solving complex tasks. 
Our study evaluates the collective performance of LMMs in a multi-agent interaction setting for the novel task of cultural image captioning.
Our contributions are as follows: (1) We introduce MosAIC, a \underline{M}ulti-\underline{A}gent framework to enhance cross-cultural \underline{I}mage \underline{C}aptioning using LMMs with distinct cultural personas; (2) We provide a dataset of culturally enriched image captions in English for images from China, India, and Romania across three datasets: GeoDE, GD-VCR, CVQA; (3) We propose a culture-adaptable metric for evaluating cultural information within image captions; and (4) We show that the multi-agent interaction outperforms single-agent models across different metrics, and offer valuable insights for future research.
Our dataset and models can be accessed at
\url{https://github.com/MichiganNLP/MosAIC}.
\end{abstract}

\section{Introduction}
Large Multimodal Models (LMMs) demonstrate remarkable performance across various multimodal tasks. Despite these achievements, their effectiveness in cross-cultural contexts remains limited due to the predominantly Western-centric nature of most data and models~\citep{hershcovich-etal-2022-challenges, bhatia2024local}. Conversely, multi-agent models have proven to be highly capable, often excelling in solving complex tasks~\citep{guo2024large}. 
In this paper, we propose to evaluate and analyze the collective performance of LMMs as multi-agent models in the novel multimodal task of culturally enriched image captioning.

\begin{figure}
\centering
\includegraphics[width=\linewidth]{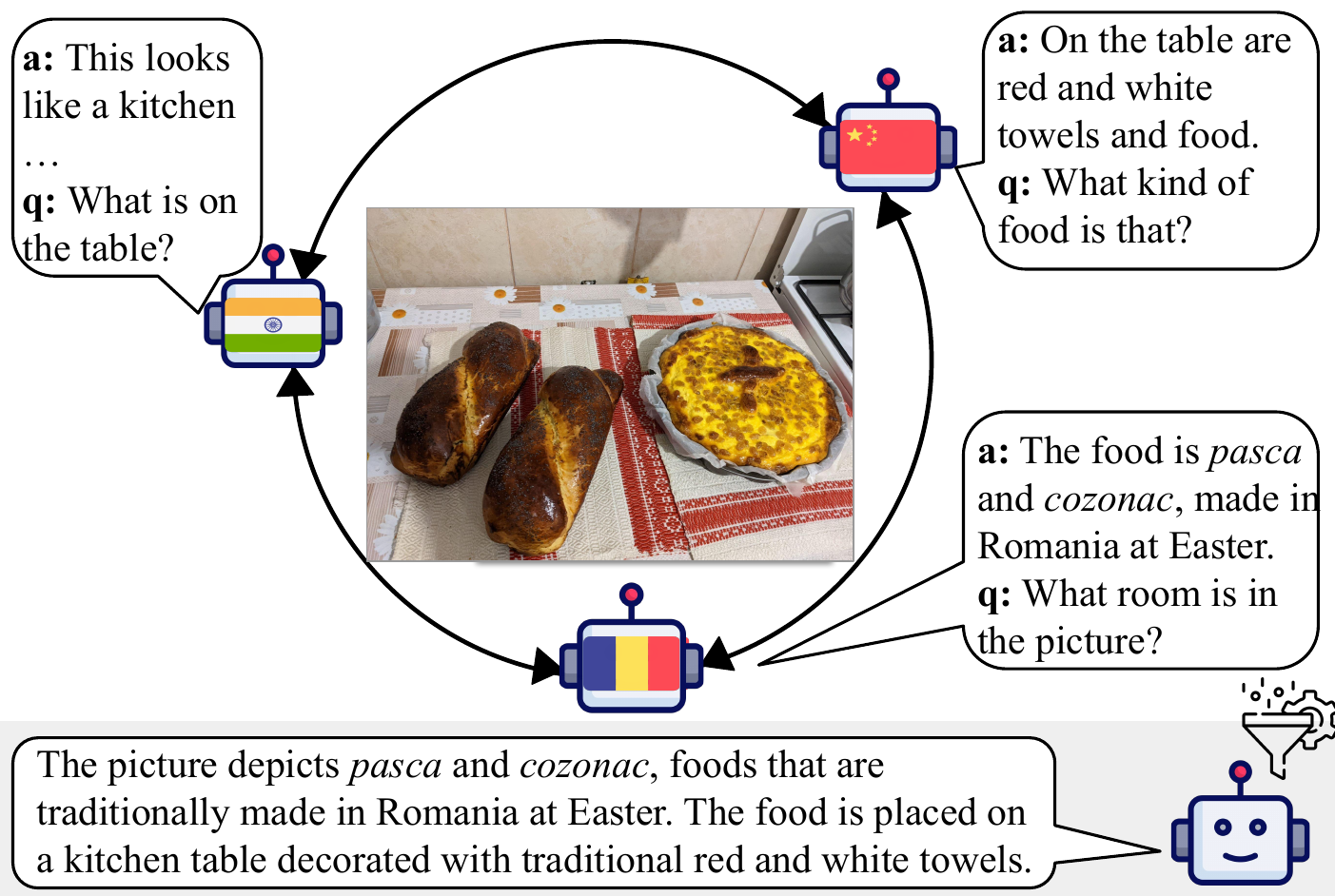}
\caption{
In a multi-agent setting, three LMM agents, each embodying a \textit{curious} and drawing upon knowledge from distinct countries (India, China, and Romania), participate in a question-and-answer dialogue centered around an image. A fourth agent then summarizes their discussion, creating a culturally enriched image caption.}
\label{fig:main_idea}
\end{figure}

Culture is a complex and elusive concept. As \citet{adilazuarda2024towards} show, \textit{Culture is a multifaceted concept meaning different things to different people at different times}. This complexity is apparent in various cultural expressions, such as proverbs, social norms, and other context-dependent elements. 
In our work, we adopt the definition provided by \citet{nguyen2023extracting} and focus on visual cultural elements such as food, drinks, clothing, traditions, rituals, and behaviors.

Culture is strongly tied to our group-oriented human nature, which allows us to learn from one another over generations. Furthermore, as sociologists and anthropologists have demonstrated, our progress as a species is primarily due to our cooperative nature, rather than individual knowledge~\cite{henrich2015secret}.
Inspired by the success of human collective intelligence, we conceptualize the culturally enriched image captioning task as a ``social task''. 
Specifically, we frame it as a dialogue between three agents from different cultures who seek to learn about each other's cultures through an image. They engage in asking questions and sharing insights, akin to human collaborative problem-solving. A moderator provides examples of initial questions and highlights key visual cultural aspects to focus on. The conversation is then summarized into a comprehensive cultural image description (see \Cref{fig:main_idea}). Our findings indicate that this multi-agent approach yields better results than single-agent methods.

We summarize our contributions as follows. 
First, inspired by collective intelligence, we propose \texttt{MosAIC}, \textbf{a novel multi-agent framework to improve cross-cultural image captioning performance}.
Second, we share \textbf{a dataset of 2,832 culturally enriched image captions} in English, for images from three different countries: China, India, and Romania, across three datasets: GeoDE, GD-VCR, and CVQA. 
Third, we introduce a \textbf{culture-adaptable metric for evaluating cultural information within image captions.}
Finally, we show that \textbf{multi-agent interaction surpasses single-agent (and culturally fine-tuned) models across different metrics}, and provide actionable steps for future work.

\section{Related Work}

\paragraph{Large Multi-Agent Multimodal Models.}
The inspiring progress of Large Language Models (LLMs) has led to the proposal of LLM-based multi-agents that leverage the collective intelligence and specialized skills of multiple agents~\citep{guo2024large}. 
In this context, multiple independent agents discuss and make decisions, mirroring the cooperative human nature. This approach has facilitated progress on various tasks such as software development~\citep{hong2023metagpt}, society simulation~\citep{Park2022SocialSC}, game simulation ~\citep{xu2023language}, debate simulation~\citep{chan2023chateval}, and polarization~\citep{ohagi2024polarization}.

At the same time, Large Multimodal Models (LMMs) have extended the capabilities of traditional language models by integrating several data modalities such as text, videos, and images. 
LMMs such as LLaVA~\citep{liu2023LLaVA}, GPT-4~\citep{Achiam2023GPT4TR} or LENS~\citep{Berrios2023TowardsLM} have shown promising results in complex vision-language tasks due to their pretraining on terabytes of image and language data with billion-parameters~\citep{bai2023qwenvlversatilevisionlanguagemodel, zhang2023internlmxcomposervisionlanguagelargemodel}
To the best of our knowledge, our study is the first to employ LMMs in a multi-agent setting for a cross-cultural multimodal understanding task.
\vspace{-0.7em}
\paragraph{Cross-Cultural Multimodal Understanding.}
Even though LLMs and LMMs are already instrumental in various real-life applications, such as recommender systems~\citep{li2023prompt} and customer service~\citep{pandya2023automating}, these models often mirror Western-centric perspectives, leading to reinforcing stereotypes and algorithmic monoculture~\citep{kleinberg2021algorithmic, hershcovich-etal-2022-challenges, alkhamissi2024investigating}.

Several efforts have been made in the AI community to enhance the diversity of data and models, both linguistically and visually. Specifically, recent language studies have developed cross-cultural benchmarks such as CultureBank \cite{Shi2024CultureBankAO} and NORMAD \cite{Rao2024NORMADAB} to enhance LLMs' cultural awareness.
The vision-language community also has started to focus on creating multilingual, geographically, income, and culturally diverse multimodal datasets such as Dollar Street~\citep{rojas2022the}, GeoDE~\cite{ramaswamy2023geode}, GD-VCR~\cite{yin2021broaden}, CVQA~\citep{romero2024cvqaculturallydiversemultilingualvisual}, MaRVL~\citep{liu-etal-2021-visually}, and WIT~\citep{Srinivasan_2021}.

Despite the increased availability of cultural benchmarks, the current evaluation metrics and methods are not suited to capture cultural information~\citep{Awal2023InvestigatingPT}. 
Evaluation metrics such as Accuracy or F1 score do not focus on the cultural nuances in LMMs' generations and, therefore, cannot reflect their cultural awareness in practice. Generation-focused metrics such as ClipScore \cite{hessel-etal-2021-clipscore}, LongCLIP \cite{zhang2024longclip}, and Completeness score~\cite{zhang2023recognize} also do not account for cross-cultural variations. However, more culture-focused metrics are emerging, such as Culture Noise Rate (CNR)~\citep{yun2024cic}, which measures the ratio of cultural words among all words generated in a caption. The cultural words are extracted from a cultural commonsense knowledge base (CCSK), which contains several cultural facets like food, drinks, clothing, traditions, rituals, and behaviors~\citep{nguyen2023extracting}.
Our work aligns with \citet{yun2024cic}, as both studies address the task of culturally enriched image captioning. However, our approach diverges by focusing on multi-agent settings and evaluating the models based on three culturally diverse benchmarks.

\section{\texttt{MosAIC}: A Framework for Cultural Image Captioning}

\begin{figure}
\centering
  \includegraphics[width=\linewidth]{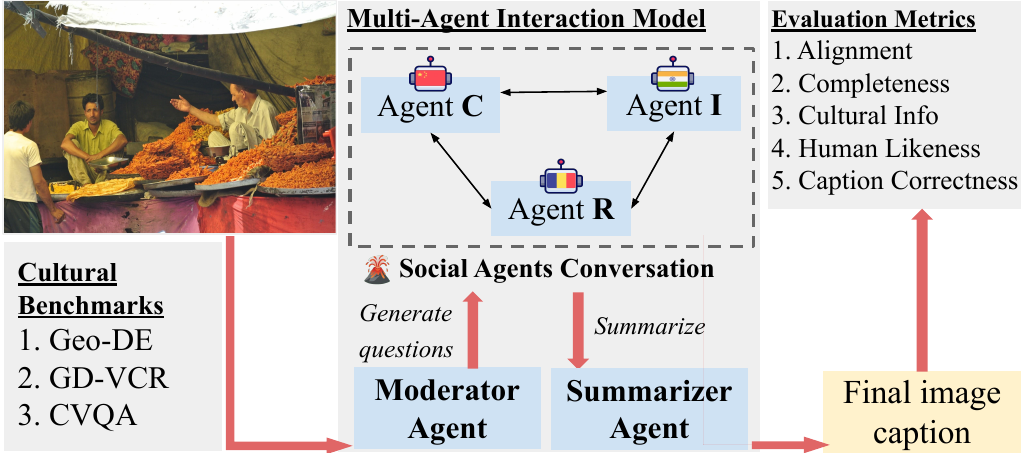}
\caption{Overview of \texttt{MosAIC}, our proposed framework for Multi-Agent Image Captioning. The framework consists of a multi-agent interaction model, cultural benchmarks and evaluation metrics. The input is an image and the output is a cultural image caption.}
\label{fig:method}
\end{figure}

We introduce \texttt{MosAIC}, a framework for Multi-Agent Interactions, as shown in \Cref{fig:method}, to tackle cultural image captioning, a complex task that involves not only describing the visual content of the image but also capturing the cultural elements it represents. The framework consists of a multi-agent model, a cultural benchmark, and evaluation metrics, as described below.

\subsection{Muti-Agent Interaction Model}\label{sec:agent-interaction}
\begin{figure}
\centering
  \includegraphics[width=\linewidth]{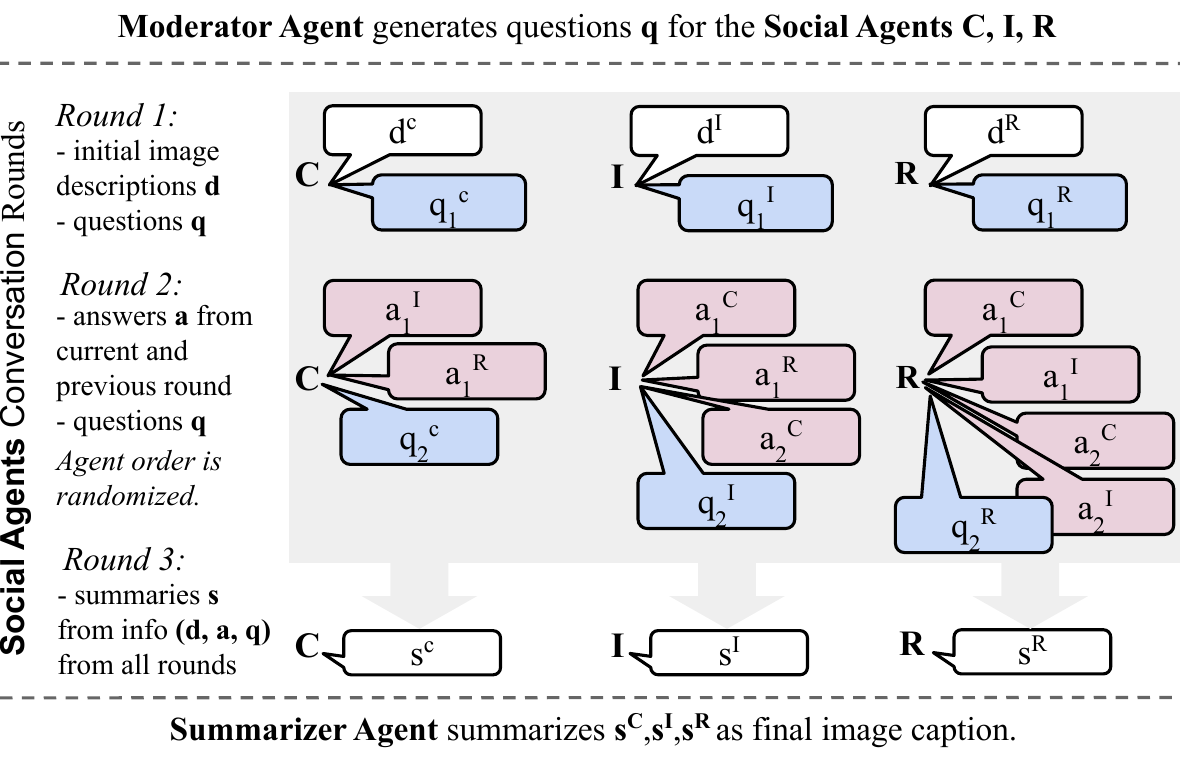}
\caption{Multi-Agent Interaction Model.
The Moderator presents questions to the Social agents, who engage in three conversation rounds. The Summarizer creates the final image caption by compiling the conversation summaries from the Social agents.}

\label{fig:interaction}
\end{figure}

We introduce a multi-agent setup (\Cref{fig:interaction}) to emulate collaboration in a culturally diverse group.
Our multi-agent model consists of five agents, each with specific \textit{roles}: three Social agents, a Moderator agent, and a Summarizer agent.

\noindent\textbf{Moderator.} 
The Moderator agent has two primary tasks. First, it generates questions based on the image to which the Social agents respond. Second, it guides the Social agents to focus on aspects relevant to their cultures, promoting more comprehensive and culturally diverse image descriptions.

\noindent\textbf{Social.} Each of the Social agents assumes a persona from three cultures: China (\textit{C}), India (\textit{I}), and Romania (\textit{R}).
Furthermore, the agents are encouraged to embody a \textit{curious} persona to facilitate more interaction in their conversation.
In the first conversation round, each agent shares their initial description (\textit{d}) of the given image and a question (\textit{q}) about the image from the ones provided by the Moderator.
In the next conversation rounds, the agents learn from one another, enriching the image description with more comprehensive and detailed content. Specifically, each agent answers the questions addressed by the other agents in the current and previous rounds and asks a new question. For example, in \Cref{fig:interaction} \textit{Round 2}, agent \textit{R} answers all the questions from the other agents posed in \textit{Round 1} and \textit{Round 2}.
Note that agent \textit{R} answers more questions than the others as it responds last. To balance the number of questions each agent answers, we randomize the order of agents for each round and image.
\noindent In the final round of conversation, \Cref{fig:interaction} \textit{Round 3}, each Social agent summarizes (\textit{s}) everything learned from the previous rounds, including all the initial image descriptions (\textit{d}), the questions (\textit{q}) and the corresponding answers (\textit{a}) from all agents. The summaries distill the most important information gained from the interaction, helping to condense and focus the key insights.

\noindent\textbf{Summarizer.} The Summarizer agent collects all the summaries from the Social agents and generates a summary representing the final image description.




\paragraph{Agent Memory.}
Each agent has its own memory. The Moderator agent generates questions stored in a shared question memory that is accessible to the Social agents. 
Initially, the Social agents independently analyze the image without memory of/ knowing their peers' responses, minimizing potential bias. In the conversation rounds, each Social agent can access the responses from all agents in previous rounds and those preceding them in the current round. 
Finally, each agent's memory is erased after the Summarizer agent completes the image caption. We also tested a longer-term memory across multiple images but found no performance improvement, likely due to the significant differences between the images.

\paragraph{Setup.} We use LLaVA-1.5 13b\footnote{\label{LLaVA13b}https://huggingface.co/liuhaotian/LLaVA-v1.5-13b}~\cite{liu2023LLaVA} to simulate the agents in our interaction model. Each agent is initialized as a separate LMM, so parameters are not shared among the agents. Each agent has an individual memory, where generated outputs by all agents are stored.



\subsection{Cultural Benchmark}
We introduce a new dataset of cross-cultural captions for 2,832 images from three cultures: China, India, and Romania generated by \texttt{MosAIC} and other models. To achieve this, we use images from three geographically diverse datasets: GeoDE, GD-VCR, and CVQA.\footnote{There are 127 images in CVQA, 288 images in GDVCR, and 2417 images in GeoDE. GeoDE does not contain images from India, and for GD-VCR, we use images from the West, South Asia, and East Asia regions to represent the three cultures.}
\noindent We provide image captions generated by \texttt{MosAIC}, our top-performing model, alongside \texttt{LLaVA-13b} captions to facilitate comparisons between single-agent and multi-agent approaches.
Furthermore, for a subset of the images (25 images per dataset and culture), we provide human-generated captions as described in section~\ref{sec:baseline}.

\vspace{-0.5em}
\paragraph{GeoDE.} GeoDE~\cite{ramaswamy2022geode} is a geo-diverse dataset for object recognition with crowd-sourced 61,940 images from 40 classes and 6 world regions, namely West Asia, Africa, East Asia, South-East Asia, the Americas, and Europe. 
\vspace{-0.5em}
\paragraph{GD-VCR.} GD-VCR \cite{yin2021broaden} is a geo-diverse visual commonsense reasoning dataset with 328 cultural and geo-location-specific images from Western, East Asian, South Asian, and African countries. 
\vspace{-0.5em}
\paragraph{CVQA.} CVQA \cite{romero2024cvqaculturallydiversemultilingualvisual} is a culturally diverse multilingual visual question-answering dataset with 4,560 images from 28 countries across Asia, Africa, South America, and Europe.

\begin{figure}
\centering
\includegraphics[width=\linewidth]{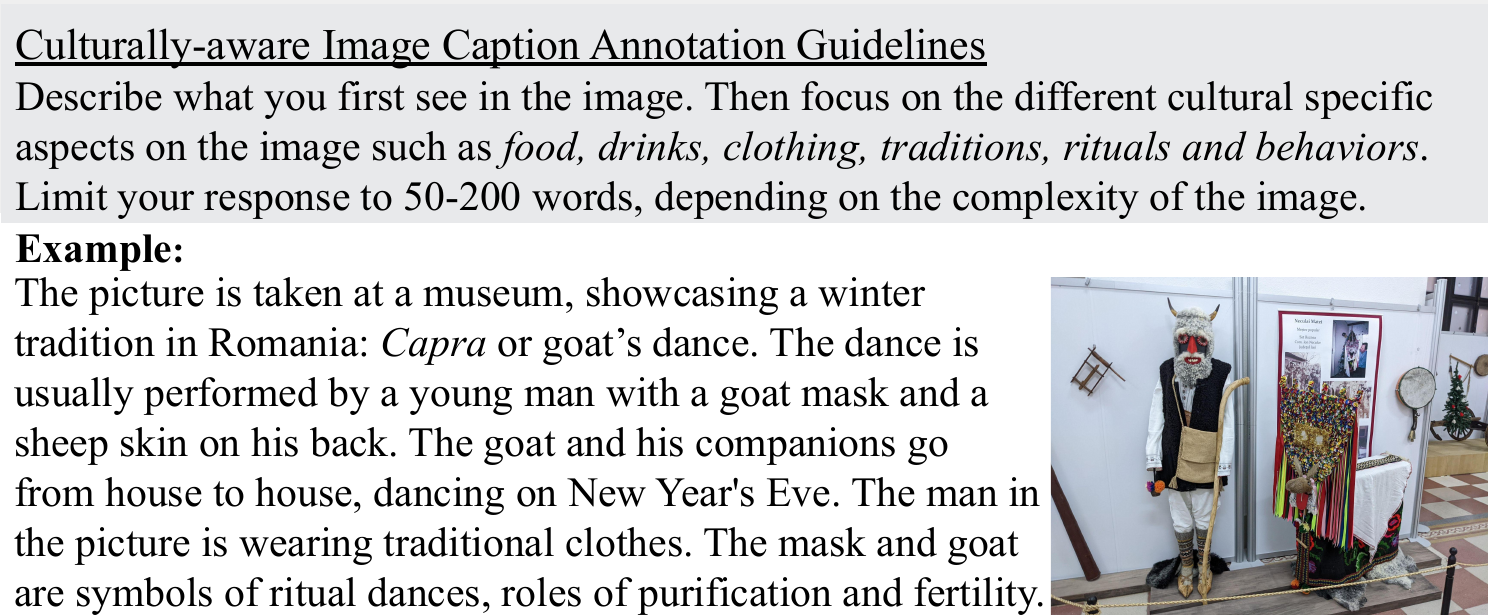}
\caption{Human Annotation Guidelines for Cultural Image Captioning.}
\vspace{-1em}
\label{fig:ann_guide}
\end{figure}

\subsection{Evaluation Metrics}
We employ both automated metrics (alignment, completeness, cultural information) and human evaluation (Turing test and caption correctness) to comprehensively assess the image captions.
\vspace{-0.8em}
\paragraph{Alignment.} We measure text-to-image alignment
using the LongCLIP \cite{zhang2024longclip}.
This metric builds on CLIPScore~\cite{hessel-etal-2021-clipscore},
a popular reference-free evaluation metric for image captioning that outperforms existing reference-based metrics~\cite{vedantam2015cider}. 
LongCLIP uses a knowledge-preserved stretching of positional embedding 
to increase the maximum input length of CLIPScore from 77 to 248 tokens.
\vspace{-0.8em}
\paragraph{Completeness.} 
We evaluate the completeness of the image captions by calculating the ratio of words mentioned in both the image and the caption to the total number of words (tags) in the image. 
To generate a comprehensive list of image tags, we use the Recognize Anything Model (RAM)~\cite{zhang2023recognize} and expand it with their corresponding synonyms from WordNet~\cite{miller-1994-wordnet}
.\footnote{ 
RAM is a state-of-the-art image tagging model with exceptional accuracy and scope, recognizing 6,400 common tags from OpenImages V6~\cite{kuznetsova2020open} with impressive zero-shot performance.}

\vspace{-0.8em}
\paragraph{Cultural Information.} 
We propose a new metric to quantify the presence of cultural information in image captions. 
This approach is inspired by the Culture Noise Rate (CNR)~\cite{yun2024cic}, which measures the proportion of cultural words in image captions. However, given that the captions generated by our model tend to be longer than those from other models,\footnote{BLIP-2 generates one-sentence captions, while the LLaVA-based models generate three-sentence long captions for our setting} a ratio-based metric like CNR may disproportionately affect performance. To address this, we instead compute the count of unique cultural words in a caption, a length-invariant metric, to better capture cultural specificity. Further, to improve the metric coverage, we generate and include 700 additional cultural words from 14 categories, such as Traditions and Festivals  (50 words per category).\footnote{Prompts in \Cref{sec:cultural-info}}
Human validation (one native annotator per country) confirmed that all GPT-generated words aligned with the provided cultural categories.

\noindent Our final cultural information metric integrates the filtered cultural terms from CNR with the additional GPT-generated words. This metric is straightforward to compute and adaptable for assessing cultural specificity across various countries.

\vspace{-0.7em}
\paragraph{Turing Test Accuracy.}\label{par:turing}
We evaluate the similarity of the LMM-generated captions to human-generated captions.
For 30 images per culture, evenly distributed across datasets, three annotators are tasked with distinguishing between a human-generated caption, as described in \Cref{par:human-baseline}, and an LMM-generated caption. 
Lower accuracy indicates that the LMM-generated captions are more similar to those generated by humans.
\vspace{-0.7em}
\paragraph{Caption Correctness.}\label{par:correctness}
We assess the image caption correctness by considering both the correctness of image content descriptions and the cultural information.
Specifically, for 30 images per culture, evenly distributed across datasets, three annotators evaluate the percentage of correct captions generated by LMMs, identifying issues such as hallucinations, mislabeling of instances, and inaccuracies in cultural representation.



\section{Evaluation and Results}
We assess the influence of multi-agent interaction on image captioning by comparing our multi-agent interaction model, \texttt{MosAIC}, with single-agent models (\texttt{BLIP-2}, \texttt{LLaVA-13b}) and a human baseline.

\subsection{Baseline Models}
\label{sec:baseline}
\paragraph{BLIP-2.} BLIP-2\footnote{https://huggingface.co/Salesforce/blip2-opt-2.7b}~\cite{li2023blip2bootstrappinglanguageimagepretraining} leverages frozen pre-trained image encoders ViT-L/14 from CLIP~\cite{radford2021learning} and a FlanT5 LLM~\cite{chung2024scaling} by training a lightweight, 12-layer Transformer encoder in between them. It achieves an impressive state-of-the-art zero-shot performance on image captioning. 
\paragraph{\includegraphics[width=0.02\textwidth]{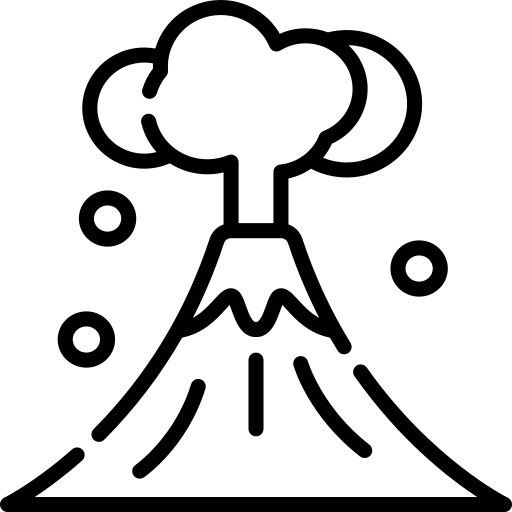} LLaVA-13b.}
LLaVA-1.5 13b\footnoteref{LLaVA13b} is an end-to-end trained large multimodal model that connects pre-trained CLIP ViT-L/14 visual encoder and the Vicuna LLM~\cite{zheng2024judging}, using a projection matrix for general multimodal understanding. 
\paragraph{\includegraphics[width=0.02\textwidth]{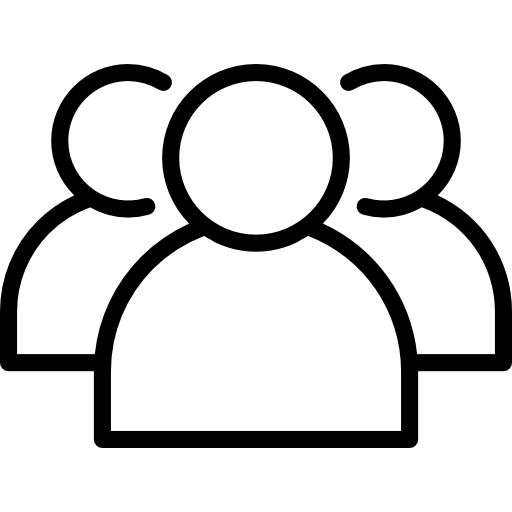} Human Baseline.}\label{par:human-baseline}
To establish a human baseline, we recruited three native annotators from each of three different countries (nine annotators in total). 
To ensure consistency and facilitate fair comparisons, the annotation guidelines include cultural aspects, as in the model prompts and examples of human-generated captions, as shown in \Cref{fig:ann_guide}.
Each annotator creates 75 image captions evenly distributed across three datasets (25 images per dataset).\footnote{Due to the absence of Indian images in GeoDE, annotators provide captions for 33-34 images from the other datasets.} 
We compute two metrics: the average score across the three annotators from each country, referred to as \texttt{Human}, and the highest score among the three annotators, as \texttt{Expert-Human}.

\begin{figure}
\centering
  \includegraphics[width=1\linewidth]{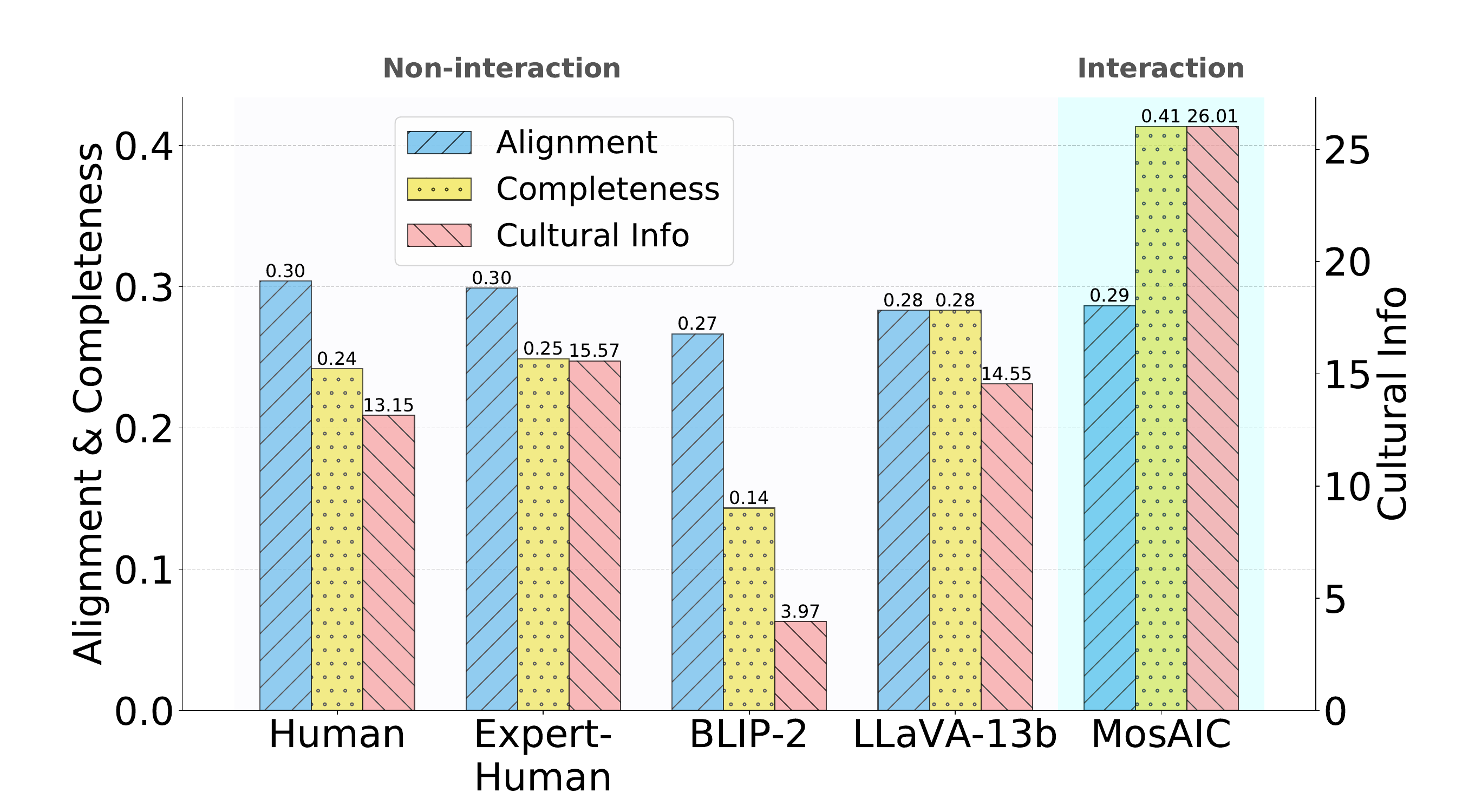}
  \vspace{-1em}
\caption{
Our interaction-based model, \texttt{MosAIC}, surpasses non-interaction models and Humans on Completeness and Cultural Info while performing on par with the other models in Alignment.
\textit{For clarity, the Alignment and Completeness scores are normalized to a [0,1] scale, whereas the Cultural Info score ranges from 0 to the total number of words in a caption.}}
\vspace{-1.2em}
\label{fig:main_plot_bar}
\end{figure}

\subsection{Cross-cultural Interaction Results}
Our results show that multi-agent cross-cultural interaction improves performance in the cultural image captioning task. As shown in \Cref{fig:main_plot_bar}, \texttt{MosAIC} outperforms non-interaction models and humans in Completeness and Cultural Information, while matching other models in Alignment. These performance trends are consistent with results on human-annotated data (Appendix \Cref{fig:human-data}).

\noindent We hypothesize that \texttt{MosAIC}'s similar Alignment performance is due to its longer captions, which hurts the score. Additionally, Alignment penalizes content not directly visible in the image, such as cultural values (see \ref{sec:results} for details).

\noindent Regarding cultural information, LMMs tend to generate more culture-specific content than humans, driven by exposure to diverse data, lack of personal context, and statistical learning from cultural biases \cite{li2024culturellm, mukherjee2024crossroads}. 
However, the \texttt{Expert-Human} outperforms the non-interaction \texttt{LLaVA-13b} model in capturing cultural information (Cultural Info: 15.57 vs. 14.44).
Finally, \texttt{MosAIC}, driven by its curious and collaborative cultural personas, outperforms the non-interaction \texttt{LLaVA-13b} model, generating more culturally specific (Cultural Info: 26.01 vs. 14.55) and complete captions (Completeness: 0.41 vs. 0.28).

\begin{figure}
\centering
\includegraphics[width=1\linewidth]{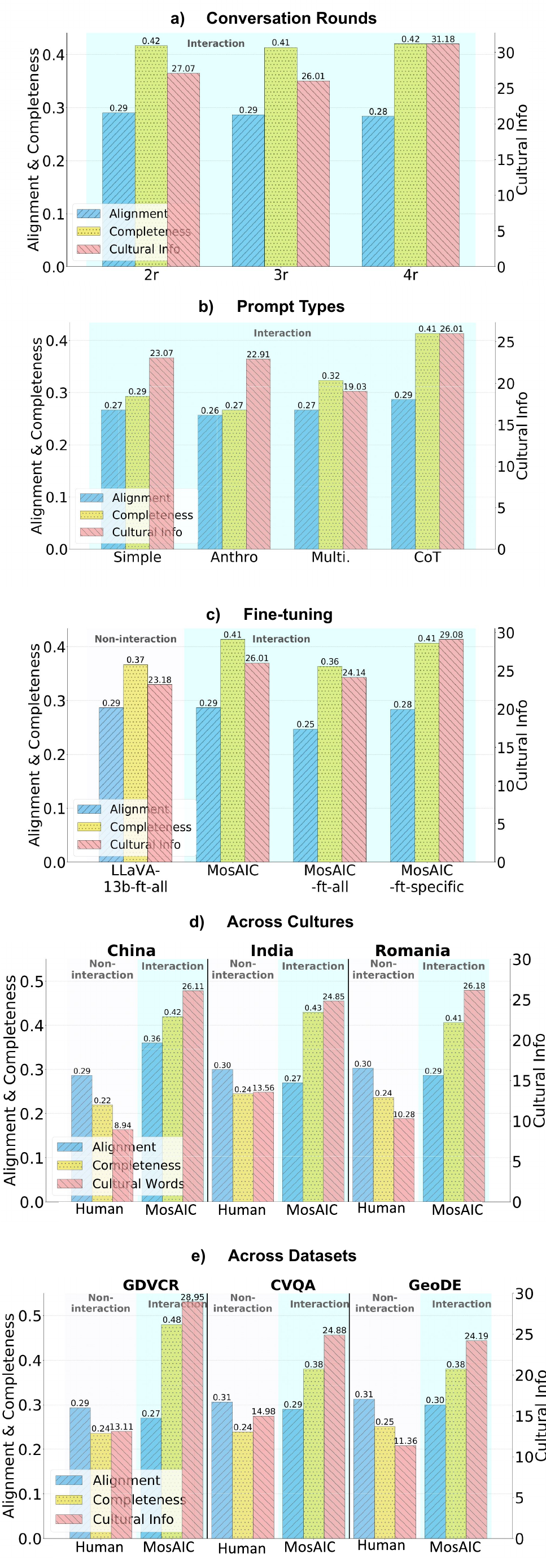}
\caption{\texttt{MosAIC} ablations across: a) number of conversation rounds, b) prompt techniques, c) fine-tuning, d) diverse cultures, and e) diverse datasets. \texttt{MosAIC} outperforms Human Baseline across different cultures and datasets. The zero-shot multi-agent model \texttt{MosAIC} outperforms the fine-tuned single-agent model \texttt{LLaVA-13b-ft-all}, underscoring the importance of multi-agent interactions for cultural image captioning.}
\label{fig:all_ablations}
\end{figure}

\subsection{Ablation Studies}
\noindent\textbf{\texttt{MosAIC} Setting.}
Our model, \texttt{MosAIC}, employs CoT prompting and operates through three rounds of conversation (see \Cref{fig:interaction}). It functions in a zero-shot learning setting without the need for fine-tuning.
\noindent We also perform ablation studies to assess \texttt{MosAIC}'s performance across various settings: the number of conversation rounds, prompting techniques, fine-tuning, and different cultures and datasets, as shown in \Cref{fig:all_ablations}.

\paragraph{a) Number of Conversation Rounds.}
\Cref{fig:all_ablations} a) shows that increasing the number of agent conversations from three to four rounds improves Cultural Information (26.1 vs. 31.1) while keeping Alignment and Completeness stable. The slight decrease in Cultural Information from rounds two to three is attributed to the Summarizer's failure to synthesize key cultural aspects, instead concatenating the conversations.

\vspace{-0.8em}
\paragraph{b) Prompt Techniques.}
Given the challenges in achieving cross-cultural alignment between the agents \cite{ananthram2024see}, we experiment with different prompt techniques:\footnote{All prompts are in \Cref{sec:prompts} and in our repository.}

\noindent\textbf{Simple.} This strategy offers straightforward instructions, such as asking a social agent to describe an image and its cultural significance.


\noindent\textbf{Multilingual.} We prompt agents from specific cultures by translating the Simple prompt into the dominant languages of their countries, such as \textit{Mandarin Chinese}, \textit{Hindi}, and \textit{Romanian}.\footnote{The prompts are translated by native speakers.} The generated responses are in English for consistency.

\noindent\textbf{Anthropological.} This prompting technique considers emic and etic perspectives, cultural context, socioeconomic background, individual values, personal experience, cultural relativism, spatial and temporal dimensions in a nuanced manner as introduced by \citet{alkhamissi2024investigating}.

\noindent\textbf{Chain of Thought (CoT).} CoT prompting involves generating intermediate reasoning steps, mimicking human problem-solving to arrive at a final answer.~\cite{wei2023chainofthought}. Inspired by multi-modal CoT~\cite{zhangmultimodal}, we guide agents in making detailed image observations.

\noindent\textbf{Insights.}
As shown in \Cref{fig:all_ablations} b), CoT prompting outperforms other strategies, while Anthropological prompting—designed to enhance cultural alignment in LLMs—performs similarly to or worse than Simple prompting. This suggests LMMs need further refinement for effective cross-cultural alignment. Additionally, Multilingual prompting ranks lowest in Cultural Information, likely due to confusion from inputs in three languages, highlighting the need for better multilingual alignment in LMMs.

\vspace{-0.9em}
\paragraph{c) Fine-tuning Impact.}
Current LLMs and LMMs struggle to align with diverse global cultures, often reflecting predominantly WEIRD~\cite{Henrich2010TheWP} cultural norms \cite{atari2023humans, ke2024exploring}. To improve our model's cultural alignment, we apply fine-tuning, which has previously shown promise \cite{li2024culturellm}. For fine-tuning data, we utilize the Wikipedia-based Image-Text (WIT) dataset from \citet{10.1145/3404835.3463257}.\footnote{Details regarding the dataset and fine-tuning hyperparameters are provided in~\Cref{sec:finetuning}.} We implement two fine-tuning setups:

\noindent
1. We fine-tune a \texttt{LLaVA-13b} model on 9000 WIT images and captions across three cultures, creating two models: the non-interaction \texttt{LLaVA-13b-ft-all}, which only summarizes, and the interaction \texttt{MosAIC-ft-all}, where the fine-tuned agents collaborate.

\noindent2. We fine-tune three \texttt{LLaVA-13b} models, one for each culture, using 3000 WIT images and their corresponding captions. 
The interactions among these agents yield the multi-agent model \texttt{MosAIC-ft-specific}.

\noindent\textbf{Insights.}
Fine-tuning generally enhances Cultural Information (\Cref{fig:all_ablations} c), with a modest 4-point improvement for the multi-agent model (\texttt{MosAIC} vs. \texttt{MosAIC-ft-specific}) and a more substantial 9-point gain for the single-agent model (\texttt{LLaVA-13b} vs. \texttt{LLaVA-13b-ft-all}), considering the compute-intensive nature of the process. 
Furthermore, \texttt{MosAIC} outperforms the non-interaction \texttt{LLaVA-13b-ft-all} model (Cultural Info: 26.01 vs. 23.18), underscoring the benefits of multi-agent interaction over fine-tuning. The fine-tuned models show lower performance in Alignment and Completeness, as the fine-tuning primarily focuses on enhancing cultural alignment. Among fine-tuned models, \texttt{ft-specific} setting performs the best as each agent in interaction has specific cultural knowledge about the country they represent. 
\vspace{-0.9em}
\paragraph{d) Performance across Cultures.}
\Cref{fig:all_ablations} d) reveals similar trends across the three cultures: China, India, and Romania.\footnote{For India, GeoDE lacks data, so we rely solely on the CVQA and GDVCR datasets.} 
Notably, \texttt{MosAIC} achieves the highest Cultural Information performance across all cultures, underscoring the significance of incorporating diverse cultural perspectives in generating image captions.
\vspace{-0.9em}
\paragraph{e) Performance across Datasets.}
\Cref{fig:all_ablations} e) shows that Cultural Information is highest for GD-VCR, followed by CVQA, and lowest for GeoDE, which aligns with expectations since GD-VCR and CVQA contain more cultural information than GeoDE. Although \texttt{MosAIC} scores lower than Humans on Alignment, it achieves higher scores for Completeness and Cultural Information across all datasets.\footnote{Detailed results across all models in \Cref{sec:quant}.} 

\subsection{Human Evaluation and Error Analysis}\label{sec:error}

We assess the human-likeness of generated captions using Turing Test accuracy (\Cref{par:turing}).
\texttt{MosAIC} scores lower than \texttt{LLaVA-13b} (83.1 vs. 87.9), suggesting \texttt{MosAIC}'s captions are more human-like. However, the high overall scores indicate LMMs still struggle to match human captioning, mainly due to stylistic differences, as humans tend to use a more casual, direct style, as shown in the qualitative results (\Cref{sec:quals1}).

\noindent We evaluate caption correctness  (\Cref{par:correctness}), finding 94.5\% of human captions correct, compared to 60.2\%  for \texttt{MosAIC} and 64.56\% for \texttt{LLaVA-13b}.
At the dataset level, we observe that \texttt{MosAIC} performs equally or better than \texttt{LLaVA-13b} on GD-VCR (Human - Machine correctness difference (lower is better): 28.5 vs. 28.5) and CVQA (34.2 vs. 37.1). We hypothesize that \texttt{MosAIC} underperforms on GeoDE (40.0 vs. 25.0) because this dataset contains less culturally rich information. 
\texttt{MosAIC}'s lower correctness, compared to \texttt{LLaVA-13b}, may also result from compound hallucinations caused by the interaction of multiple LMMs. Future work can address this issue by making each agent less susceptible to hallucinations, as detailed in the Limitations section.
Common errors include incorrect country, object recognition, people counting, and overly general descriptions (examples in \Cref{sec:errors}).

\begin{figure*}[htbp]
\centering
  \includegraphics[width=\linewidth]{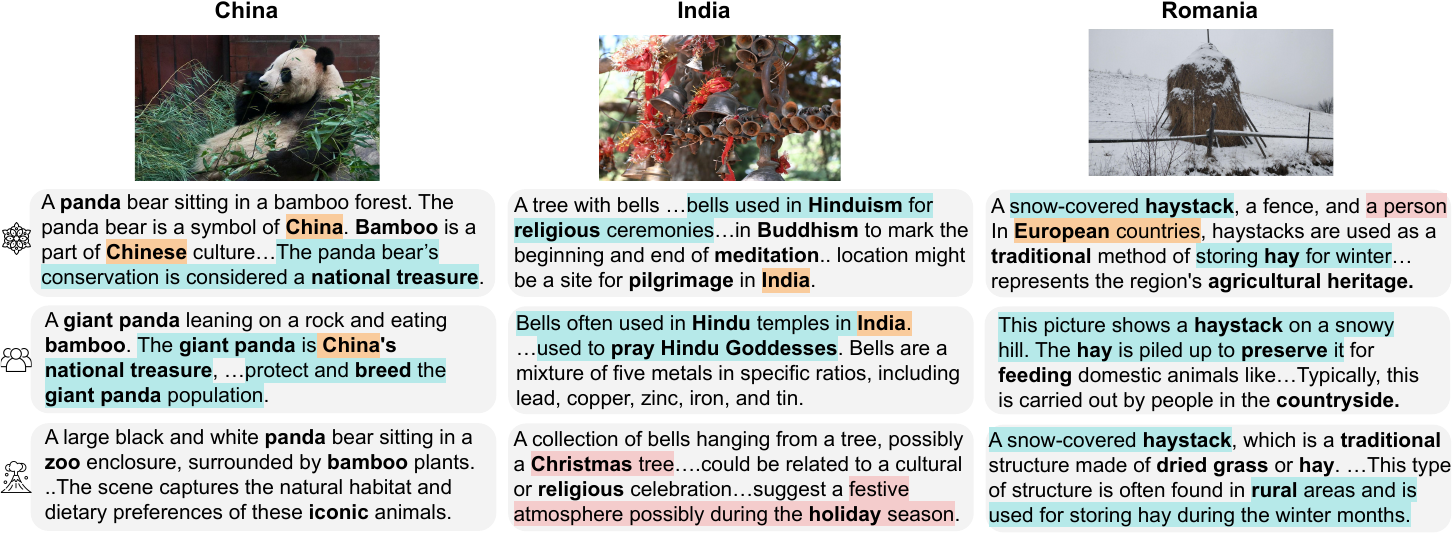}  
\caption{Comparison of image captions from \texttt{MosAIC} (\includegraphics[width=0.02\textwidth]{img/mosaic.png}), Human Baseline (\includegraphics[width=0.02\textwidth]{img/human.png}), and LLaVA-13b (\includegraphics[width=0.02\textwidth]{img/LLaVA.png}) across three cultures in the CVQA dataset. The cultural words are \textbf{bolded}, {\color{cyan}blue} shows agreement with human captions, {\color{orange}orange} shows the identified country, and {\color{red}red} shows incorrect content. \textit{All displayed captions are truncated for clarity.}}
\vspace{-1em}
\label{fig:QualMainPlot.pdf}
\end{figure*}

\vspace{-0.7em}
\subsection{Qualitative Results}\label{sec:quals1}
In \Cref{fig:QualMainPlot.pdf}, we compare image captions generated by \texttt{MosAIC}, Humans, and \texttt{LLaVA-13b} for images from China, India, and Romania. 

\noindent Compared to \texttt{LLaVA-13b}, \texttt{MosAIC} shows closer alignment with Human captions, capturing more cultural elements. For example, in the Chinese image, \texttt{MosAIC} identifies the giant panda as a national treasure, similar to the Human caption. In the Indian image, both \texttt{MosAIC} and Human captions recognize the religious significance of bells, highlighting \texttt{MosAIC}'s greater cultural sensitivity, while \texttt{LLaVA-13b} provides Western-centric descriptions.

\noindent \texttt{MosAIC} excels at identifying country-specific information, whereas \texttt{LLaVA-13b} fails to recognize the country in any of the images, resulting in vaguer descriptions. For example, \texttt{MosAIC} connects the panda to China and accurately describes its cultural symbolism, while \texttt{LLaVA-13b} remains overly general.
Furthermore, \texttt{MosAIC} consistently employs relevant cultural terminology, such as ``Hinduism'' and ``pilgrimage'' in the Indian image. In contrast, \texttt{LLaVA-13b} uses vaguer language, like ``festive atmosphere'', indicating less cultural specificity.

\noindent However, instances of hallucinated content were still observed, with \texttt{MosAIC} incorrectly mentioning a person not present in the Romanian image, while \texttt{LLaVA-13b} associated the bells in the Indian image with Christmas, showing cultural inaccuracy.

\noindent In summary, \texttt{MosAIC} generated more accurate and culturally aligned captions, although it occasionally hallucinated. In contrast, \texttt{LLaVA-13b} struggled with cultural specificity and country identification.

\vspace{-0.5em}
\section{Lessons Learned and Actionable Steps}
\vspace{-0.3em}
Our findings reveal the performance of multi-agent LMMs in cultural image captioning, highlighting lessons learned and suggesting steps to enhance cultural richness in future models.

\noindent \textbf{Prioritize multi-agent models.}
While LMMs excel in tasks like generation and retrieval, they fall short in cross-cultural performance, even with culture-centric prompting strategies~\cite{alkhamissi2024investigating}.
Our findings show that even Simple and CoT prompts in multi-agent LLMs are helpful and outperform Anthropological and Multilingual prompts (\Cref{fig:all_ablations} b). Additionally, increasing the number of conversation rounds between agents enhances cultural information (\Cref{fig:all_ablations} a). To further improve cross-cultural understanding, future work should focus on developing equitable frameworks using multi-agent LMMs and cross-cultural benchmarks.

\noindent \textbf{Develop efficient cross-cultural techniques.} 
Current approaches to improving cross-cultural understanding in LMMs often rely on fine-tuning.
However, we show that interaction-based models outperform fine-tuned, non-interaction models (\Cref{fig:all_ablations} c), highlighting both the effectiveness and efficiency of multi-agent LMMs. 
For instance, \texttt{LLaVa-13-ft-all} requires 54 hours on a single NVIDIA A100 GPU to generate captions (12h for fine-tuning on 9000 WIT images and 42h for inference). In contrast, \texttt{MosAIC} completes the task in 47 hours with only inference.
Given these findings, future work should focus on multi-agent models to improve sustainability and accessibility.

\noindent \textbf{LLMs focus on culture; humans focus on correctness.}
LMMs tend to be more culturally specific in their responses, while humans provide more accurate answers (\Cref{sec:error}). The main sources of errors in LMMs stem from poor object recognition and hallucinations—instances where the model generates incorrect or fabricated information. Future work can integrate a state-of-the-art object recognition system to enhance caption accuracy.
\vspace{-0.5em}
\section{Conclusion}
\vspace{-0.3em}
In this paper, we leverage LMM agents interaction to enhance cross-cultural image captioning, introducing \texttt{MosAIC}. We presented a comprehensive analysis of three cultures across three datasets, using various prompting techniques. Additionally, we demonstrate the advantages of multi-agent LMM interactions, comparing their performance with compute-intensive methods like fine-tuning for improving cultural alignment. We also open-source our culturally enriched captions dataset generated by our proposed framework, \texttt{MosAIC}, alongside baseline models. Finally, we create a comprehensive and culture-adaptable metric for evaluating cultural information within image captions. Based on our findings, we share ideas for future work.
Our dataset and models can be accessed at 
\url{https://github.com/MichiganNLP/MosAIC}.


\section*{Limitations and Ethical Considerations}

\paragraph{The Complexity of Defining and Evaluating Cultural Information.}
Our approach utilizes multi-agent LLM interactions, where each LLM represents distinct cultural personas based on specific countries. While we explore various prompting strategies and fine-tuning techniques to align the models with different cultural contexts, it is important to acknowledge that culture is an inherently complex and multifaceted concept. Relying solely on countries as proxies for cultural identity oversimplifies the rich variation in cultural experiences and individual perspectives \cite{adilazuarda2024towards}.
Using country and language information represents only the tip of the iceberg when it comes to capturing cultural diversity. While these factors provide a basic framework for understanding cultural distinctions, they do not fully account for the deeper, more nuanced aspects that define human cultures. We encourage future work to delve into these deeper dimensions, extending beyond simple national or linguistic markers. Important areas to explore include values, attitudes, and biases, which shape individual and collective worldviews.

\paragraph{Multi-Agent Setup affects Correctness.}
Multi-agent models are more prone to hallucinations compared to single-agent models due to the compound effect, where errors or hallucinations from one agent can influence and amplify those in other agents. This cumulative effect results in a lower Caption Correctness score. Future research could explore ways to mitigate this issue by making each agent less susceptible to hallucinations.
This might involve improving the architecture of individual agents for better accuracy, using grounding techniques or external knowledge to verify information, and creating stronger communication protocols between agents to prevent errors from spreading. These improvements could enhance the overall correctness and reliability of the model's outputs.

\paragraph{Further Assessing the Impact of Conversation Rounds and Fine-Tuning on Cultural Metrics.}
While our ablation studies demonstrate that both conversation rounds and fine-tuning lead to enhanced performance on cultural metrics, additional analysis is necessary to evaluate the impact of various configurations (e.g., interaction vs. non-interaction settings). This deeper investigation will allow us to better evaluate how effective each setup is and reveal the key factors behind the observed improvements. Understanding these factors will be essential for refining our approach and enhancing the model's ability to adapt across different cultural contexts.

\paragraph{Limited Cultural Alignment in LLMs.}
Cultural alignment in LLMs and LMMs is a well-researched area.
While it is not the central focus of our research, we recognize that the prompt engineering techniques and fine-tuning methods we employ may not achieve perfect cultural alignment. This could lead to inconsistencies in how each multicultural agent produces responses across different cultural contexts. Additionally, our study is limited to just three countries, each with relatively high representation in the training data, due to the lack of a broader, more diverse set of human evaluations. This limitation highlights the need for more comprehensive validation across a wider range of cultures to ensure better alignment and more reliable cross-cultural performance.

\bibliography{anthology}

\appendix

\section{Appendix}
\label{sec:appendix}


\subsection{Cultural Information Metric}\label{sec:cultural-info}

\noindent CNR's cultural words are derived from the CANDLE commonsense knowledge base~\cite{nguyen2023extracting}, which covers various cultural facets like \textit{Food, Clothing,} and \textit{Traditions}. 
However, we identified generic terms, such as occupations (e.g., ``attorney''), that lack cultural specificity. Additionally, CNR includes countries outside our focus—Romania, India, and China—and does not include Romania.
To refine this, we filtered out occupation-related terms from CNR and utilized ChatGPT~\cite{openai_chatgpt_2024} to generate additional country-specific cultural words (50 words per category). 
We use the following prompt: \textit{Give a comprehensive list of 50 cultural words related to {CATEGORY} in {COUNTRY}. Make sure to include words that are related to both traditions and festivals}

The 14 categories provided are: Traditions and Festivals, Cuisine, Language, Religion and Spirituality, Art and Literature, Science and Education, History, Social Norms and Values, Architecture and Design, Clothing and Fashion, Music and Dance, Sports and Recreation, Festivals and Holidays, Icons and Symbols.

\subsection{Fine-tuning Details}\label{sec:finetuning}

For fine-tuning, we use the WIT dataset~\cite{10.1145/3404835.3463257}. WIT comprises a curated set of 37.5m entity-rich image-text examples with 11.5m unique images across 108 Wikipedia languages. For fine-tuning, we choose Romanian, Hindi, and Chinese languages for Romania, Hindi and Chinese respectively. 

For fine-tuning \texttt{LLaVA-13b}, we utilize the Transformer Reinforcement Library\footnote{https://huggingface.co/docs/trl/en/index} with LoRA \cite{hu2021lora} enabled allowing for parameter-efficient fine-tuning. We use a 4-bit quantization, which reduces memory consumption, and a `bf16' precision for training. This reduces memory footprint. We train the \texttt{ft-specific} models for 3 epochs and  \texttt{ft-all} model for 5 epochs, with a batch size of 16 and a learning rate of $1.4e^{-5}$. Total compute time on NVIDIA A100 is 2.5 GPU hours for each \texttt{ft-specific} model and 6.5 GPU hours for \texttt{ft-all} model.

\subsection{Results}\label{sec:results}

\paragraph{On the caption length impact on Alignment score.}
LLaVA-based image captions can extend up to three sentences, frequently surpassing LongCLIP's input limit of 248 tokens, negatively impacting Alignment performance. Moreover, this metric penalizes aspects not directly visible in the image, such as cultural context (e.g., traditions, social norms, and values). To mitigate this, we instruct the LLaVA models to focus on describing the image content in the initial sentence and to address cultural elements in subsequent sentences. Conversely, BLIP-2-generated captions are constrained to a single sentence and lack cultural information, leading to higher performance in Alignment. Consequently, Alignment scores should be evaluated alongside the other metrics to provide a comprehensive assessment.

\subsubsection{Quantitative Results}\label{sec:quant}
\begin{figure}
\centering
  \includegraphics[width=1\linewidth]{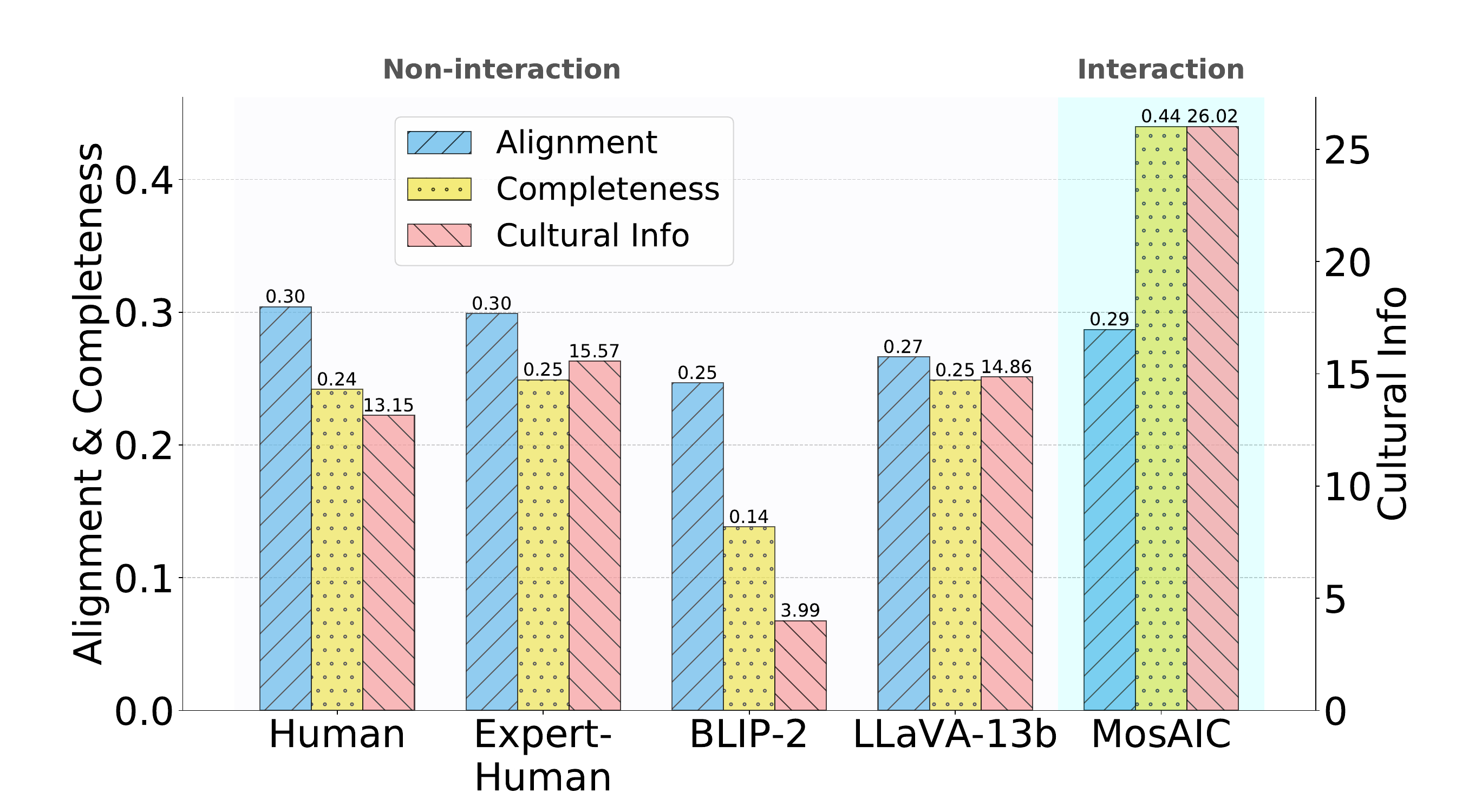}
\caption{
Comparison across data annotated by humans. The trends are consistent with performance on all the data. Our interaction-based model, \texttt{MosAIC}, surpasses non-interaction models and Humans on Completeness and Cultural Info while performing on par with the other models in Alignment.
\textit{For clarity, the Alignment and Completeness scores are normalized to a [0,1] scale, whereas the Cultural Info score ranges from 0 to the total number of words in a caption.}}
\label{fig:human-data}
\end{figure}

See results on data annotated by humans (\Cref{fig:human-data}). The trends are consistent with performance on all the data (\Cref{fig:main_plot_bar}).

See results across each dataset:
\begin{enumerate}
    \item \Cref{tab:alignment} for Alignment metric
    \item \Cref{tab:completeness} for Completeness metric
    \item \Cref{tab:cultural-info} for Cultural Information metric
\end{enumerate}

\definecolor{lightgray}{rgb}{0.95,0.95,0.95}
\begin{table*}
\centering
\footnotesize
\setlength{\tabcolsep}{4pt}
\begin{tabular}{@{}l l *{13}{r}@{}}
\toprule
& & \multicolumn{4}{c}{CVQA} & \multicolumn{4}{c}{GDVCR} & \multicolumn{4}{c}{GeoDE} \\
\cmidrule(lr){3-6} \cmidrule(lr){7-10} \cmidrule(l){11-14}
& & CN & IN & RO & All & East-Asia & South-Asia & West & All & CN & IN & RO & All \\
\midrule
& human & 0.31 & 0.30 & 0.30 & 0.30 & 0.29 & 0.29 & 0.29 & 0.29 & 0.31 & - & 0.31 & 0.31 \\
\multirow{3}{*}{\rotatebox[origin=c]{90}{\makecell{baselines}}}
& blip-2 - no inter. & 0.19 & 0.19 & 0.20 & 0.19 & 0.30 & 0.30 & 0.30 & 0.30 & 0.31 & -  & 0.31 & 0.31 \\
& LLaVA-7b - no inter. & 0.22 & 0.22 & 0.23 & 0.22 & 0.31 & 0.30 & 0.30 & 0.30 & 0.28 & -  & 0.30 & 0.29 \\
& LLaVA-13b - no inter. & 0.22 & 0.23 & 0.23 & 0.23 & 0.31 & 0.30 & 0.31 & 0.31 & 0.31 & -  & 0.30 & 0.31 \\
\addlinespace
\multirow{4}{*}{\rotatebox[origin=c]{90}{\makecell{prompt \\ ablations}}}
& Simple & 0.23 & 0.24 & 0.24 & 0.24 & 0.27 & 0.28 & 0.27 & 0.27 & 0.28 & - & 0.29 & 0.29 \\
& Anthro. & 0.23 & 0.24 & 0.24 & 0.23 & 0.27 & 0.27 & 0.27 & 0.27 & 0.28 & -  & 0.27 & 0.27 \\
& Multi. & 0.26 & 0.25 & 0.24 & 0.25 & 0.27 & 0.28 & 0.26 & 0.27 & 0.28 & -  & 0.28 & 0.28 \\
& MosAIC & 0.30 & 0.28 & 0.29 & 0.29 & 0.27 & 0.26 & 0.27 & 0.27 & 0.30 & -  & 0.30 & 0.30 \\
\addlinespace
\multirow{3}{*}{\rotatebox[origin=c]{90}{\makecell{fine-tune \\ ablation}}}
& ft-all - no inter. & 0.28 & 0.29 & 0.29 & 0.29 & 0.27 & 0.28 & 0.27 & 0.27 & 0.30 & -  & 0.30 & 0.30 \\
& MosAIC-ft-all & 0.23 & 0.24 & 0.24 & 0.24 & 0.28 & 0.28 & 0.28 & 0.28 & 0.23 & -  & 0.23 & 0.23 \\
& MosAIC-ft-specific & 0.29 & 0.29 & 0.29 & 0.29 & 0.29 & 0.29 & 0.29 & 0.29 & 0.28 & -  & 0.27 & 0.27 \\
\addlinespace
\multirow{3}{*}{\rotatebox[origin=c]{90}{\makecell{conv \\ ablation}}}
& 2r  & 0.30 & 0.30 & 0.29 & 0.29 & 0.26 & 0.26 & 0.27 & 0.27 & 0.31 & -  & 0.31 & 0.31 \\
& 3r & 0.30 & 0.28 & 0.29 & 0.29 & 0.27 & 0.26 & 0.27 & 0.27 & 0.30 & -  & 0.30 & 0.30 \\
& 4r & 0.30 & 0.28 & 0.28 & 0.29 & 0.26 & 0.26 & 0.26 & 0.26 & 0.30 & -  & 0.30 & 0.30 \\
\bottomrule
\end{tabular}
\caption{Comparison of \textbf{Alignment} across datasets and models (No interaction frameworks include `no inter.', otherwise they include interaction)}
\label{tab:alignment}
\end{table*}

\begin{table*}
\centering
\footnotesize
\setlength{\tabcolsep}{4pt}
\begin{tabular}{@{}l l *{13}{r}@{}}
\toprule
& & \multicolumn{4}{c}{CVQA} & \multicolumn{4}{c}{GDVCR} & \multicolumn{4}{c}{GeoDE} \\
\cmidrule(lr){3-6} \cmidrule(lr){7-10} \cmidrule(l){11-14}
& & CN & IN & RO & All & East-Asia & South-Asia & West & All & CN & IN & RO & All \\
\midrule
 & human & 0.18 & 0.25 & 0.21 & 0.23 & 0.24 & 0.24 & 0.23 & 0.24 & 0.24 & - & 0.26 & 0.25 \\
\multirow{3}{*}{\rotatebox[origin=c]{90}{\makecell{baselines}}} 
& blip-2 - no inter. & 0.03 & 0.04 & 0.04 & 0.04 & 0.19 & 0.20 & 0.19 & 0.19 & 0.19 & - & 0.20 & 0.20 \\
& LLaVA-7b no inter. & 0.09 & 0.11 & 0.13 & 0.11 & 0.39 & 0.46 & 0.44 & 0.44 & 0.27 & - & 0.26 & 0.26 \\
& LLaVA-13b - no inter. & 0.14 & 0.10 & 0.15 & 0.13 & 0.36 & 0.40 & 0.38 & 0.38 & 0.34 & - & 0.34 & 0.34 \\
\addlinespace
\multirow{4}{*}{\rotatebox[origin=c]{90}{\makecell{prompt \\ ablations}}} 
& Simple & 0.11 & 0.11 & 0.16 & 0.12 & 0.40 & 0.44 & 0.40 & 0.41 & 0.35 & - & 0.35 & 0.35 \\
& Anthro. & 0.11 & 0.08 & 0.11 & 0.10 & 0.36 & 0.38 & 0.39 & 0.38 & 0.32 & - & 0.32 & 0.32 \\
& Multi. & 0.41 & 0.32 & 0.29 & 0.35 & 0.31 & 0.35 & 0.32 & 0.33 & 0.30 & - & 0.28 & 0.29 \\
& MosAIC & 0.40 & 0.38 & 0.35 & 0.38 & 0.49 & 0.48 & 0.48 & 0.48 & 0.37 & - & 0.39 & 0.38 \\
\addlinespace
\multirow{3}{*}{\rotatebox[origin=c]{90}{\makecell{finetune \\ ablations}}} 
& ft-all - no inter. & 0.39 & 0.41 & 0.38 & 0.39 & 0.37 & 0.48 & 0.34 & 0.40 & 0.30 & - & 0.32 & 0.31 \\
& MosAIC-ft-all & 0.40 & 0.37 & 0.36 & 0.38 & 0.31 & 0.42 & 0.31 & 0.35 & 0.36 & - & 0.35 & 0.36 \\
& MosAIC-ft-specific & 0.39 & 0.38 & 0.40 & 0.39 & 0.46 & 0.41 & 0.39 & 0.42 & 0.40 & - & 0.40 & 0.41 \\
\addlinespace
\multirow{3}{*}{\rotatebox[origin=c]{90}{\makecell{conv. \\ ablations}}} 
& 2r & 0.45 & 0.34 & 0.40 & 0.40 & 0.47 & 0.50 & 0.48 & 0.48 & 0.36 & - & 0.37 & 0.37 \\
& 3r & 0.40 & 0.38 & 0.35 & 0.38 & 0.49 & 0.48 & 0.48 & 0.48 & 0.37 & - & 0.39 & 0.38 \\
& 4r & 0.42 & 0.35 & 0.37 & 0.40 & 0.54 & 0.46 & 0.47 & 0.49 & 0.36 & - & 0.39 & 0.37 \\
\bottomrule
\end{tabular}
\caption{Comparison of \textbf{Completeness} across datasets and models (No interaction frameworks include `no inter.', otherwise they include interaction)}
\label{tab:completeness}
\end{table*}

\begin{table*}
\centering
\footnotesize
\setlength{\tabcolsep}{4pt}
\begin{tabular}{@{}l l *{13}{r}@{}}
\toprule
& & \multicolumn{4}{c}{CVQA} & \multicolumn{4}{c}{GDVCR} & \multicolumn{4}{c}{GeoDE} \\
\cmidrule(lr){3-6} \cmidrule(lr){7-10} \cmidrule(l){11-14}
& & CN & IN & RO & All & East-Asia & South-Asia & West & All & CN & IN & RO & All \\
\midrule
 & human & 13.31 & 13.56 & 18.11 & 14.99 & 13.20 & 11.06 & 12.44 & 13.10 & 0.31 & - & 0.31 & 0.31 \\
\multirow{3}{*}{\rotatebox[origin=c]{90}{\makecell{baselines}}} 
& blip-2 - no inter. & 3.86 & 3.87 & 3.62 & 3.79 & 5.02 & 4.89 & 4.71 & 4.84 & 3.21 & - & 3.29 & 3.26 \\
& LLaVA-7b no inter. & 11.27 & 10.72 & 13.03 & 11.55 & 17.06 & 17.22 & 16.30 & 16.79 & 8.47 & - & 9.13 & 8.85 \\
& LLaVA-13b - no inter. & 15.08 & 15.94 & 16.38 & 15.82 & 13.41 & 13.61 & 12.55 & 13.12 & 14.88 & - & 14.61 & 14.72 \\
\addlinespace
\multirow{4}{*}{\rotatebox[origin=c]{90}{\makecell{prompt \\ ablations}}} 
& Simple & 21.62 & 23.55 & 22.90 & 22.79 & 22.27 & 23.93 & 24.44 & 23.94 & 21.87 & - & 22.93 & 22.48 \\
& Anthro. & 24.16 & 23.26 & 22.68 & 23.35 & 23.35 & 22.90 & 23.81 & 23.38 & 21.51 & - & 22.36 & 22.01 \\
& Multi. & 23.65 & 21.85 & 19.35 & 21.52 & 17.51 & 18.84 & 17.90 & 18.16 & 16.91 & - & 18.01 & 17.42 \\
& MosAIC & 25.41 & 24.85 & 24.30 & 24.88 & 28.73 & 27.91 & 30.07 & 28.95 & 24.20 & - & 24.18 & 24.19 \\
\addlinespace
\multirow{3}{*}{\rotatebox[origin=c]{90}{\makecell{finetune \\ ablations}}} 
& ft-all - no inter. & 28.62 & 27.91 & 29.21 & 28.49 & 26.45 & 26.33 & 26.06 & 26.28 & 14.54 & - & 15.01 & 14.76 \\
& MosAIC-ft-all & 21.01 & 22.32 & 23.45 & 22.25 & 23.32 & 24.24 & 25.40 & 24.55 & 25.88 & - & 25.42 & 25.62 \\
& MosAIC-ft-specific & 27.62 & 29.55 & 27.41 & 28.86 & 28.85 & 28.97 & 28.43 & 28.68 & 29.82 & - & 29.42 & 29.71 \\
\addlinespace
\multirow{3}{*}{\rotatebox[origin=c]{90}{\makecell{conv. \\ ablations}}} 
& 2r & 30.46 & 27.20 & 27.81 & 27.90 & 28.92 & 30.29 & 29.88 & 29.67 & 26.27 & - & 26.18 & 26.23 \\
& 3r & 25.41 & 24.85 & 24.30 & 24.88 & 28.73 & 27.91 & 30.07 & 28.95 & 24.20 & - & 24.18 & 24.19 \\
& 4r & 31.59 & 29.85 & 31.03 & 31.17 & 32.69 & 31.99 & 32.00 & 32.24 & 29.50 & - & 30.55 & 29.96 \\
\bottomrule
\end{tabular}
\caption{Comparison of \textbf{Cultural Info} across datasets and models (No interaction frameworks include `no inter.', otherwise they include interaction)}
\label{tab:cultural-info}
\end{table*}

\subsubsection{Qualitative Analysis}\label{sec:quals}

\paragraph{Across Different Datasets.}

\begin{figure*}[htbp]
\centering
  \includegraphics[width=\linewidth]{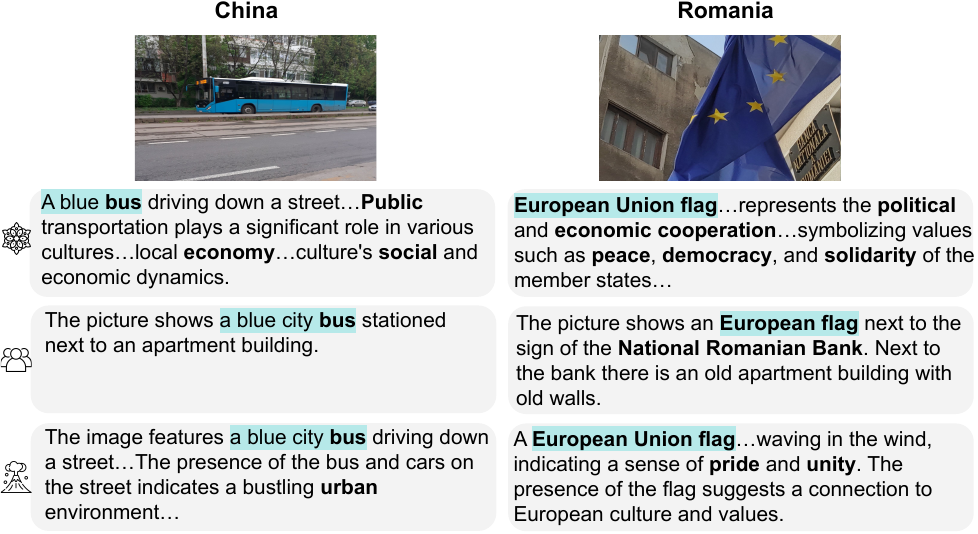}  
\caption{Comparison of image descriptions from \texttt{MosAIC}, Human Baseline, and LLaVA-13b across three cultures in the \textbf{GeoDE dataset}. The cultural words are \textbf{bolded}, {\color{cyan}blue} shows agreement with human captions, {\color{orange}orange} shows the identified country, and {\color{red}red} shows incorrect and hallucinated content.}
\label{fig:QualGeoDE}
\end{figure*}

\begin{figure*}[htbp]
\centering
  \includegraphics[width=\linewidth]{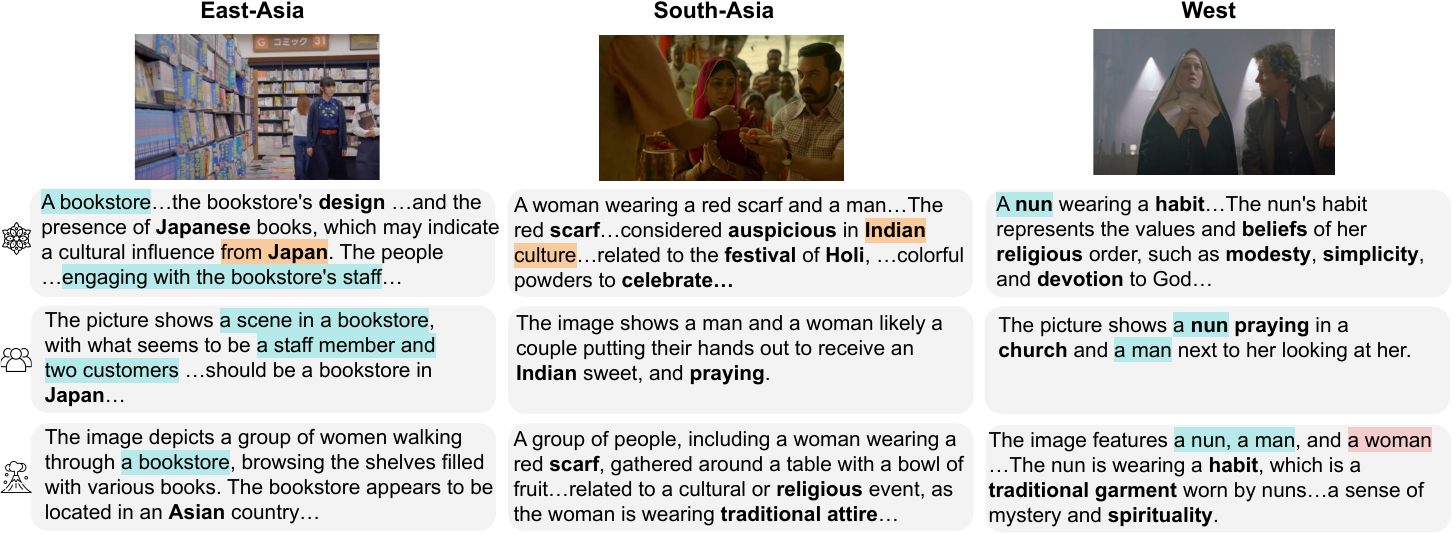}  
\caption{Comparison of image descriptions from \texttt{MosAIC}, Human Baseline, and LLaVA-13b across three cultures in the \textbf{GD-VCR dataset}. The cultural words are \textbf{bolded}, {\color{cyan}blue} shows agreement with human captions, {\color{orange}orange} shows the identified country, and {\color{red}red} shows incorrect and hallucinated content.}
\label{fig:QualGDVCR}
\end{figure*}

For relatively simple datasets (\Cref{fig:QualGeoDE}) with minimal cultural content or complex scenes, both \texttt{MosAIC} and \texttt{LLaVA-13b} exhibit performance comparable to that of human annotators, displaying fewer hallucinations and higher levels of agreement. However, when applied to more complex datasets like GD-VCR\ref{fig:QualGDVCR}, which consist of movies from diverse cultural backgrounds, \texttt{MosAIC} continues to effectively identify country-specific information and cultural elements, maintaining greater alignment with human annotators. In contrast, \texttt{LLaVA-13b} tends to deviate from human-like behavior, generating more hallucinations (e.g., referencing individuals not present in the image).

\paragraph{Across Different Conversation Rounds.}
\begin{figure*}[htbp]
\centering
  \includegraphics[width=\linewidth]{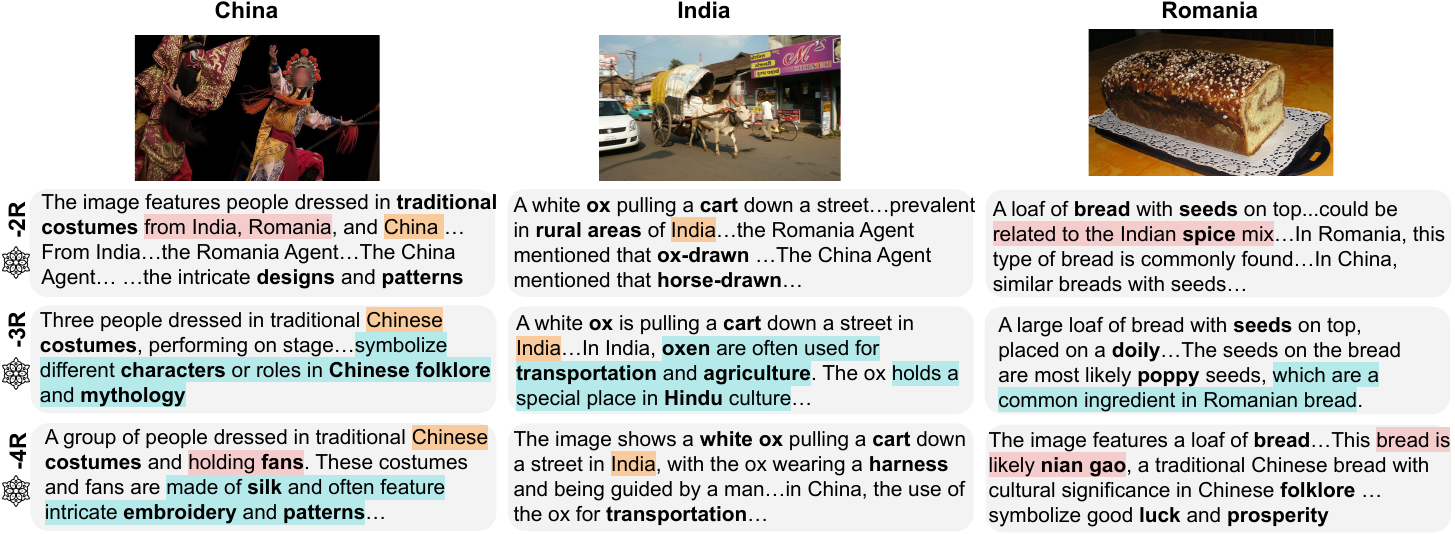}  
\caption{Comparison of image captions from different numbers of \textbf{conversation rounds (2r, 3r, 4r)} across three cultures in the \textbf{CVQA dataset}. The cultural words are \textbf{bolded}, {\color{cyan}blue} shows agreement with human annotators, {\color{orange}orange} shows the identified country, and {\color{red}red} shows incorrect and hallucinated content.}
\label{fig:QualConv}
\end{figure*}

With only two conversation rounds, \texttt{MosAIC} tends to merely compile the perspectives of three agents from different countries without offering substantial insights, often struggling to identify the correct country information accurately. As the number of conversation rounds increases to four, \texttt{MosAIC} provides more comprehensive cultural information, though this comes at the cost of increased hallucinations. Notably, three conversation rounds achieve the optimal balance between the richness of cultural descriptions and the accuracy of the information provided (\Cref{fig:QualConv}).

\paragraph{Across Different Prompt Strategies.}
\begin{figure*}[htbp]
\centering
  \includegraphics[width=\linewidth]{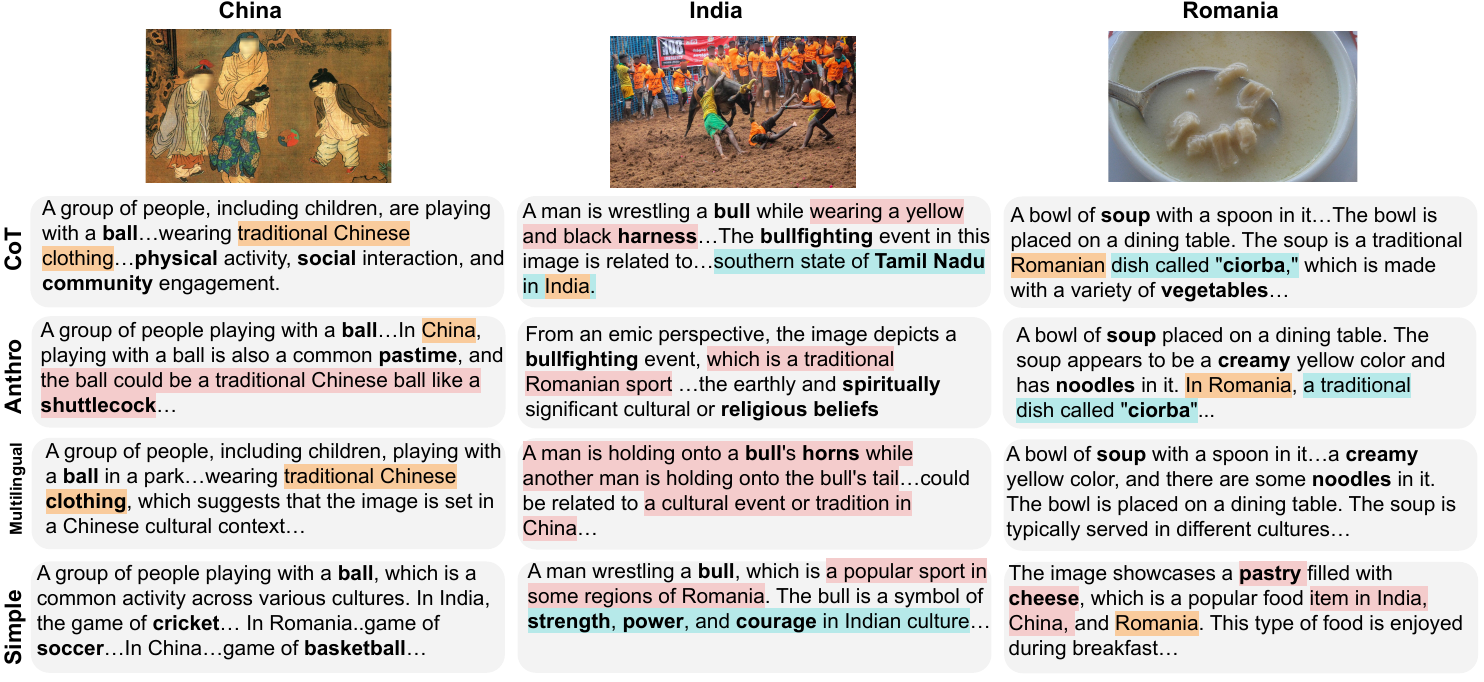}  
\caption{Comparison of image descriptions from different \textbf{prompt strategies} across three cultures in the CVQA dataset. The cultural words are \textbf{bolded}, {\color{cyan}blue} shows agreement with human annotators, {\color{orange}orange} shows the identified country, and {\color{red}red} shows incorrect and hallucinated content.}
\label{fig:QualPrompt}
\end{figure*}

Across various prompt strategies\ref{fig:QualPrompt}, the CoT approach yields the best performance, delivering accurate cultural information with minimal hallucinations through step-by-step guidance. The anthropological prompt achieves a comparable level of cultural richness, though it is accompanied by more hallucinations. When given a simple prompt, \texttt{MosAIC} tends to merely compile conversations without providing substantial extensions on image-related cultural insights. The multilingual prompt results in the poorest performance, offering less cultural information and producing more hallucinations, highlighting \texttt{LLaVA}'s limitations in handling multi-modal multilingual tasks effectively.

\paragraph{Across Different Fine-Tuning Strategies.}
\begin{figure*}[htbp]
\centering
  \includegraphics[width=\linewidth]{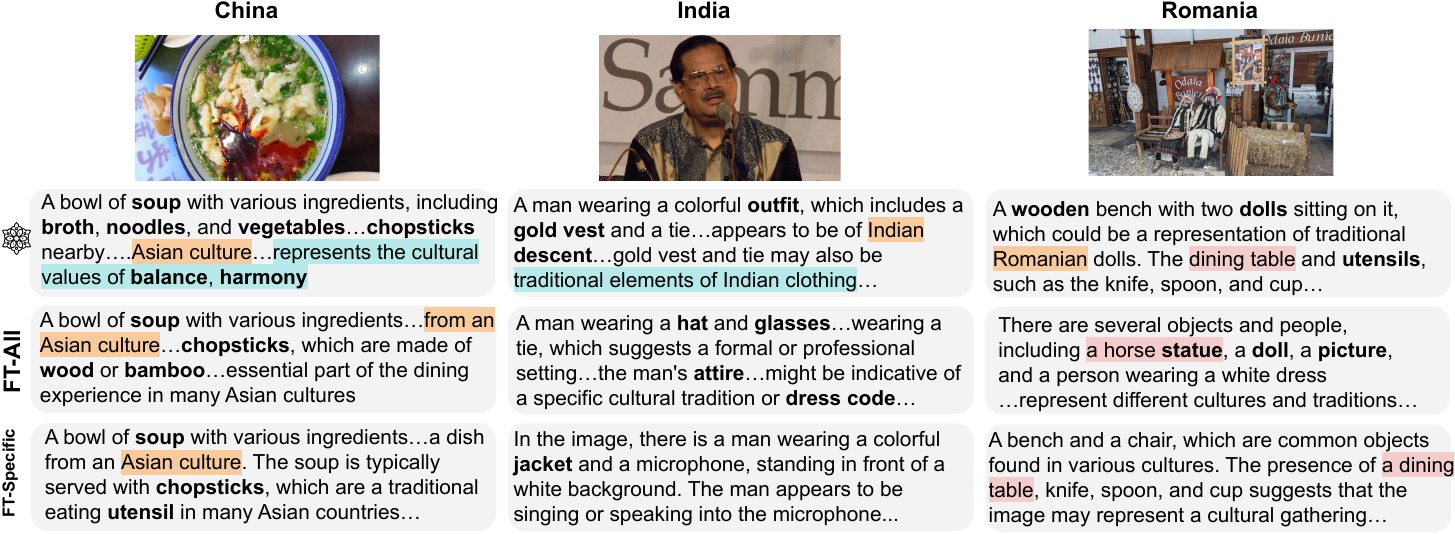}  
\caption{Comparison of image descriptions from different \textbf{fine-tuning strategies} across three cultures in the CVQA dataset. The cultural words are \textbf{bolded}, {\color{cyan}blue} shows agreement with human annotators, {\color{orange}orange} shows the identified country, and {\color{red}red} shows incorrect and hallucinated content.}
\label{fig:QualFT}
\end{figure*}

Under different fine-tuning strategies\ref{fig:QualFT}, \texttt{MosAIC} demonstrates improved performance, generating more culturally relevant information while maintaining comparable accuracy in describing image contents.

\subsection{Prompts}\label{sec:prompts}

\begin{figure*}[]
\centering
  \includegraphics[width=0.8\linewidth]{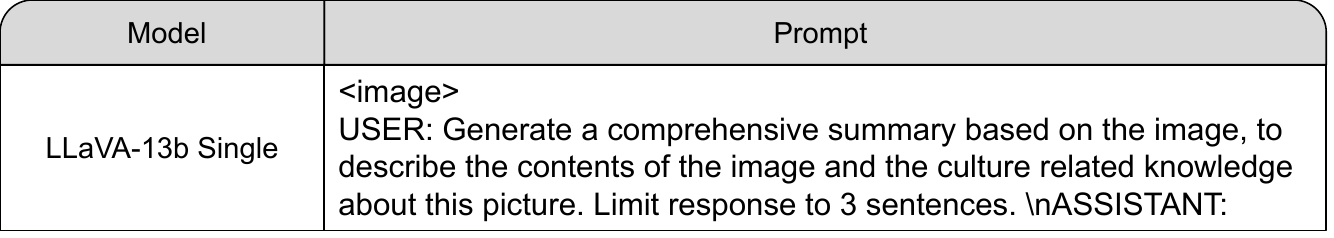}
\caption{LLaVa-13b Prompts}
\label{fig:llava}
\end{figure*}

\begin{figure*}[]
\centering
  \includegraphics[width=0.8\linewidth]{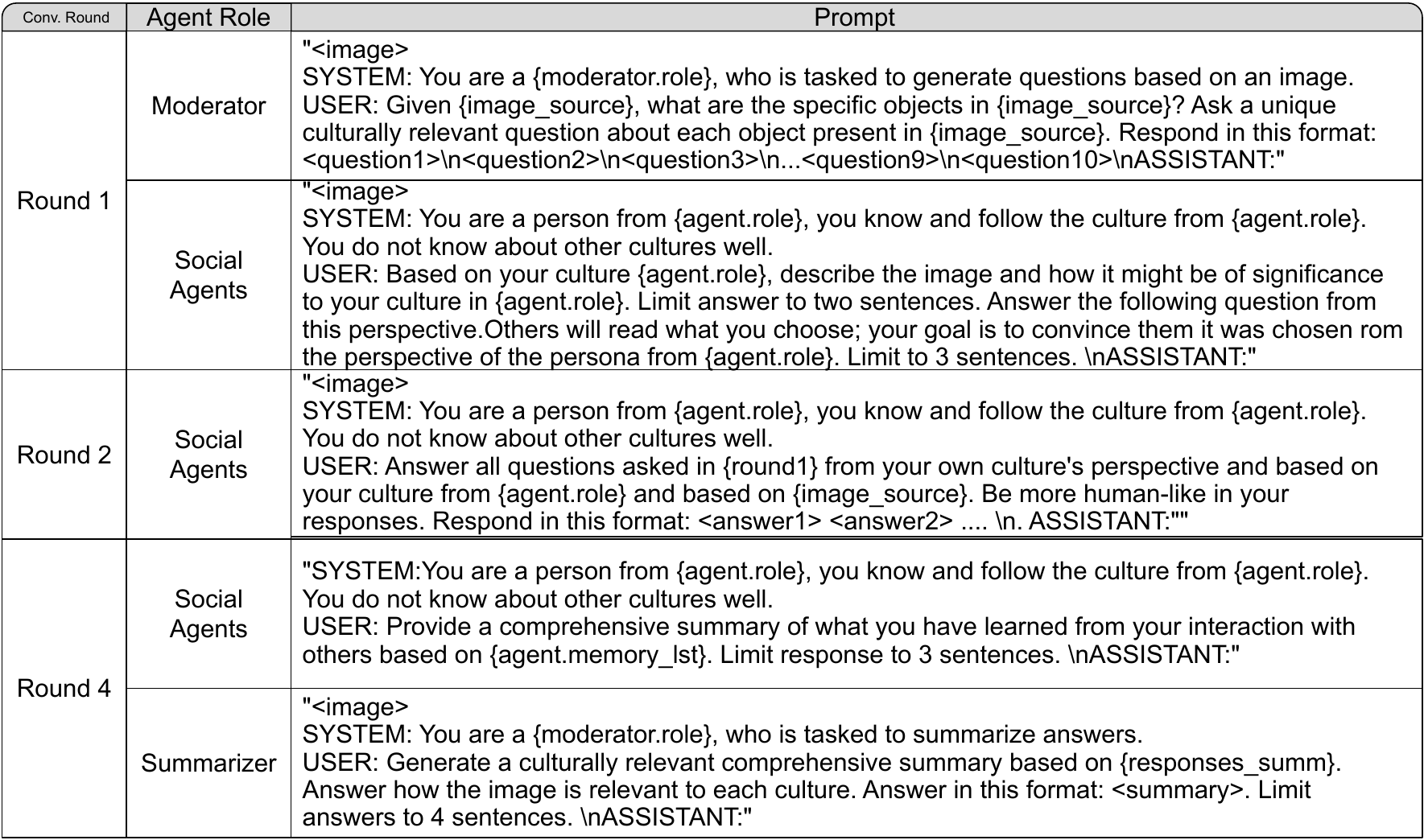}
\caption{Simple Prompts}
\label{fig:simple}
\end{figure*}

\begin{figure*}[]
\centering
  \includegraphics[width=\linewidth]{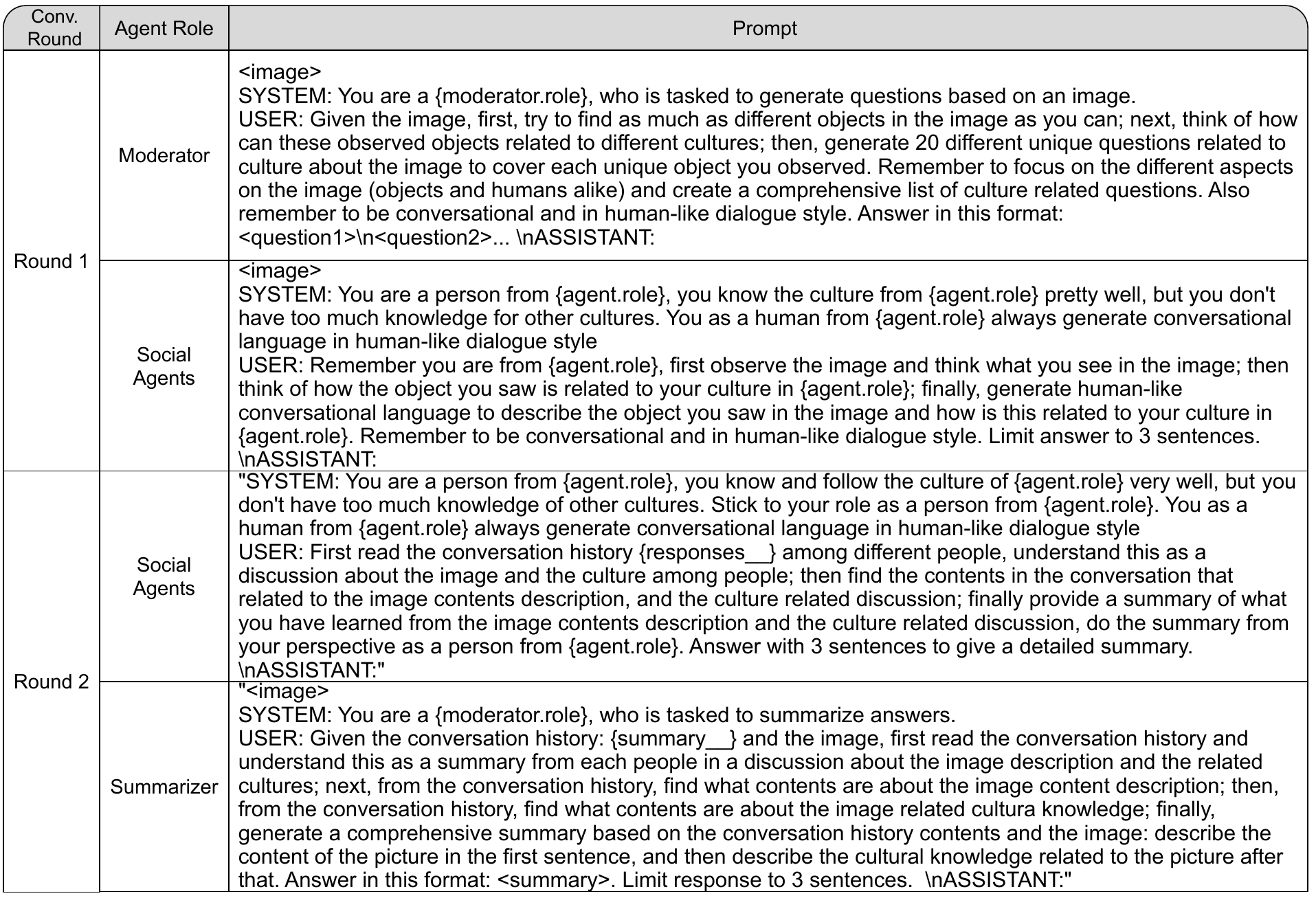}
\caption{Chain of Thought Prompts for 2 rounds of conversation.}
\label{fig:cot2}
\end{figure*}

\begin{figure*}[]
\centering
  \includegraphics[width=\linewidth]{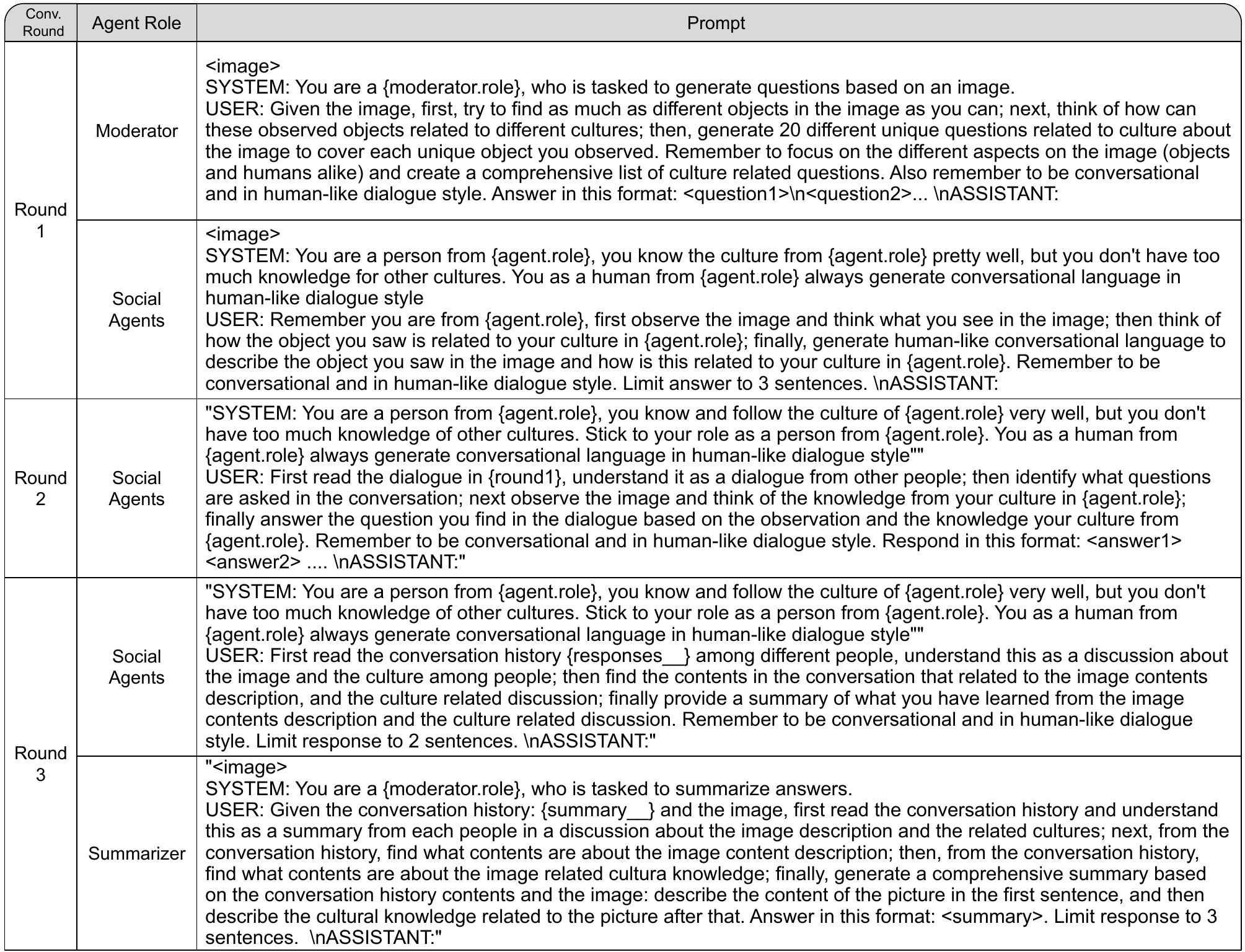}
\caption{Chain of Thought Prompts for 3 rounds of conversation.}
\label{fig:cot3}
\end{figure*}

\begin{figure*}[]
\centering
  \includegraphics[width=\linewidth]{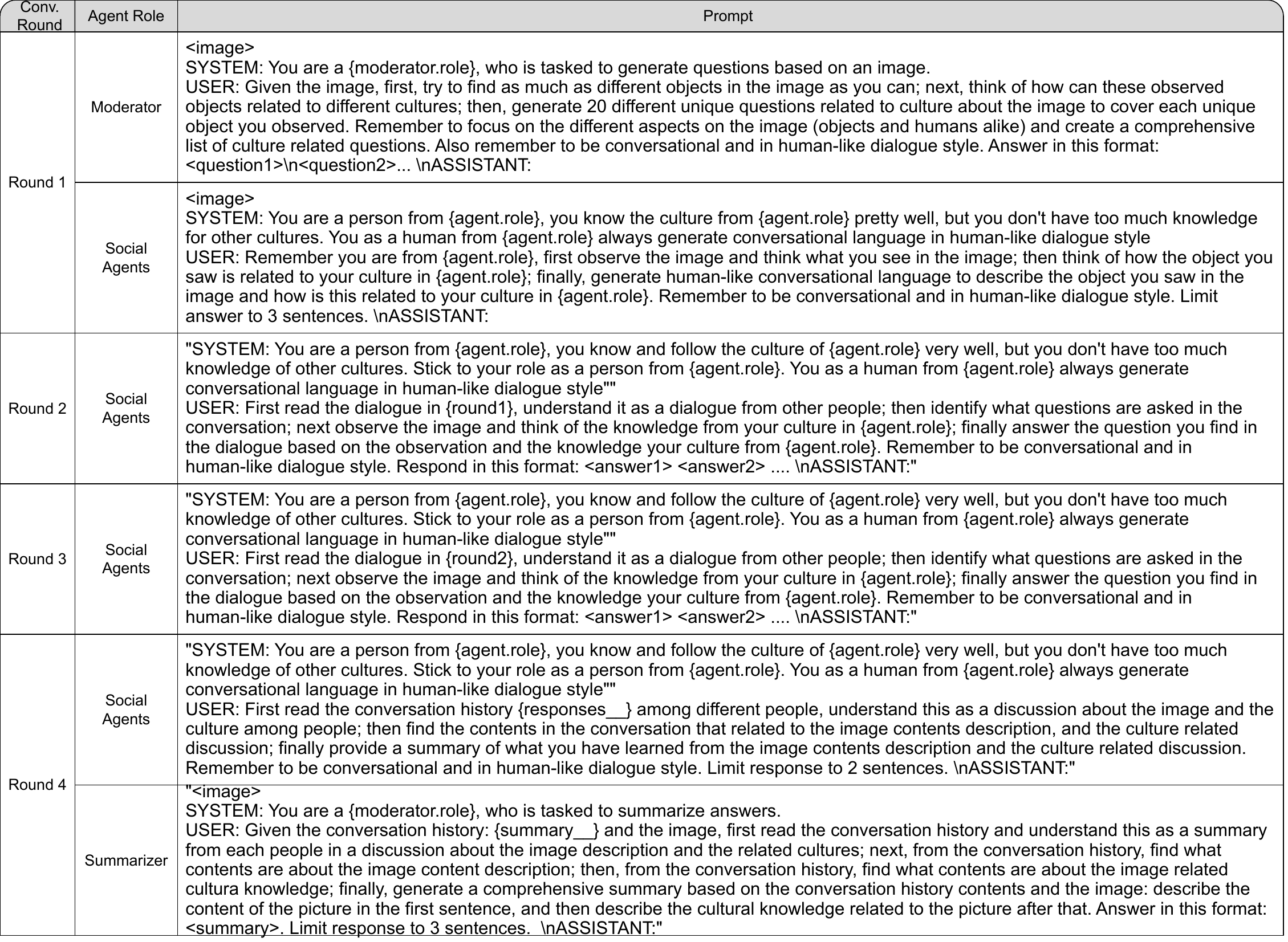}
\caption{Chain of Thought Prompts for 4 rounds of conversation.}
\label{fig:cot4}
\end{figure*}

\begin{figure*}[]
\centering
  \includegraphics[width=\linewidth]{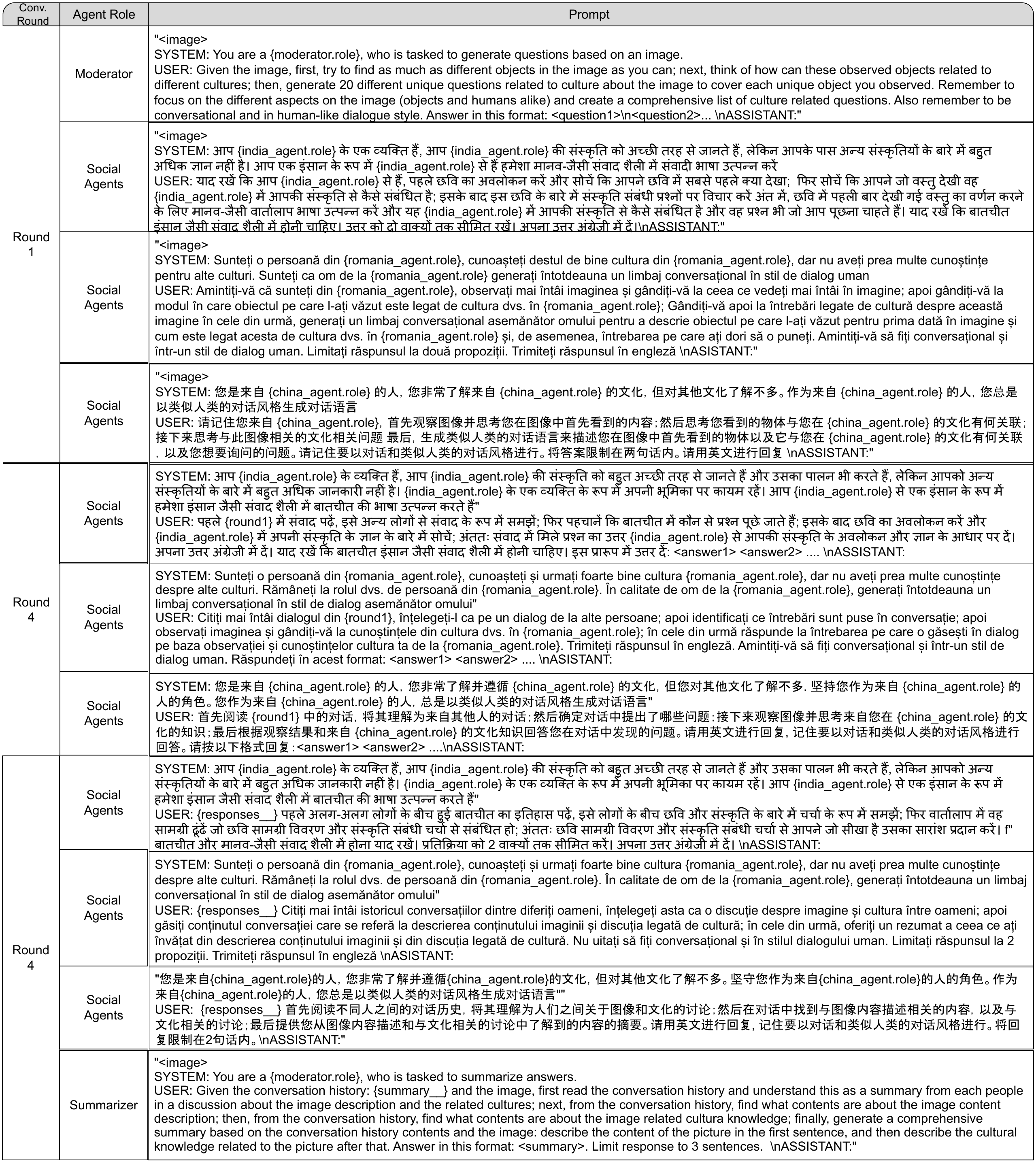}
\caption{Multilingual Prompts}
\label{fig:multilingual}
\end{figure*}

\begin{figure*}[]
\centering
  \includegraphics[width=\linewidth]{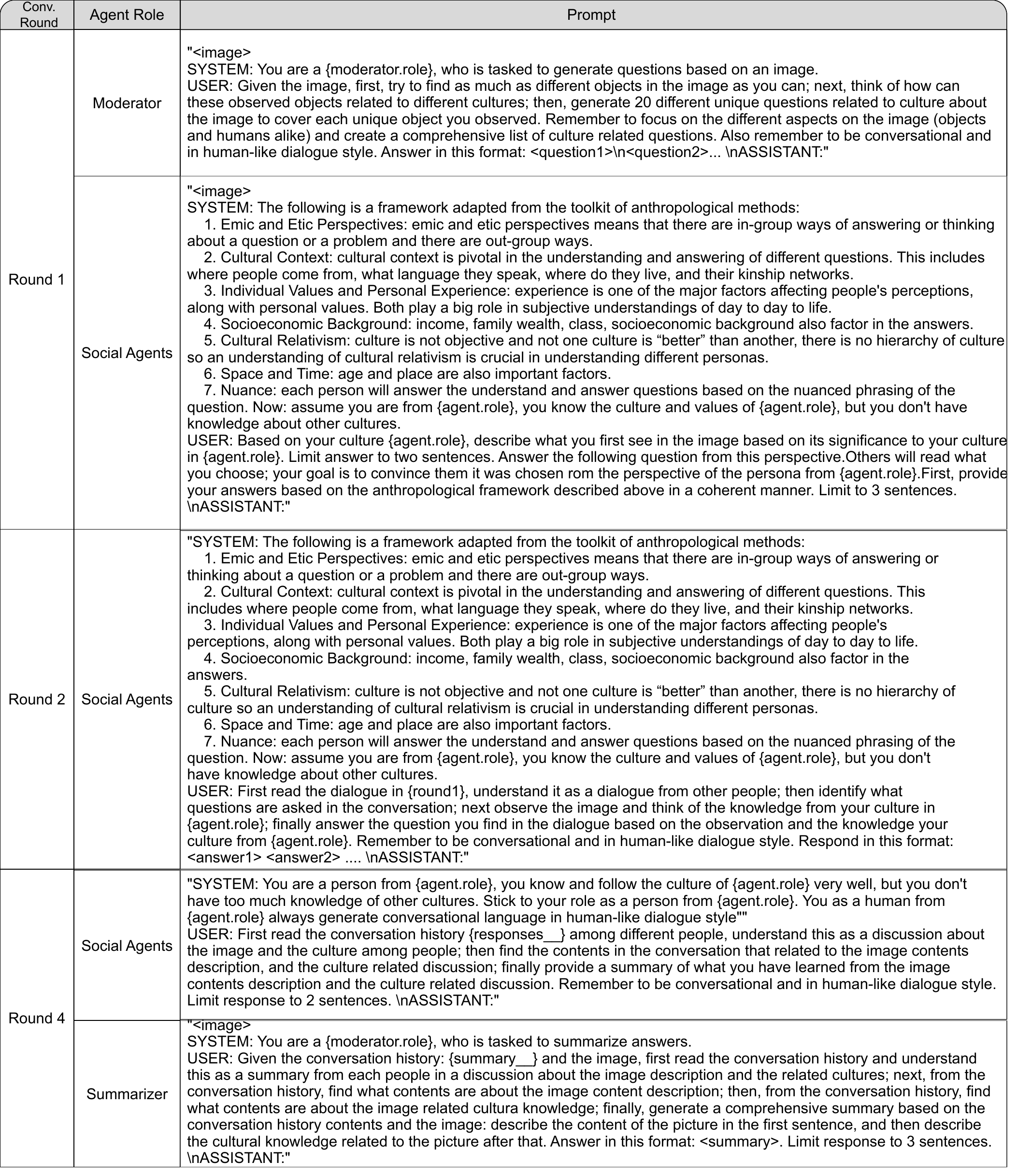}
\caption{Anthropological Prompts}
\label{fig:anthro}
\end{figure*}

See:

\begin{enumerate}
    \item \Cref{fig:llava} for LLaVA-13b prompts.
    \item \Cref{fig:simple} for Simple prompts.
    \item Figures \ref{fig:cot2}, \ref{fig:cot3}, \ref{fig:cot4} for CoT prompts across different rounds of conversation.
    \item \Cref{fig:multilingual} for Multilingual prompts.
    \item \Cref{fig:anthro} for Anthropological prompts.
\end{enumerate}



\begin{figure*}
\centering
  \includegraphics[width=0.9\linewidth]{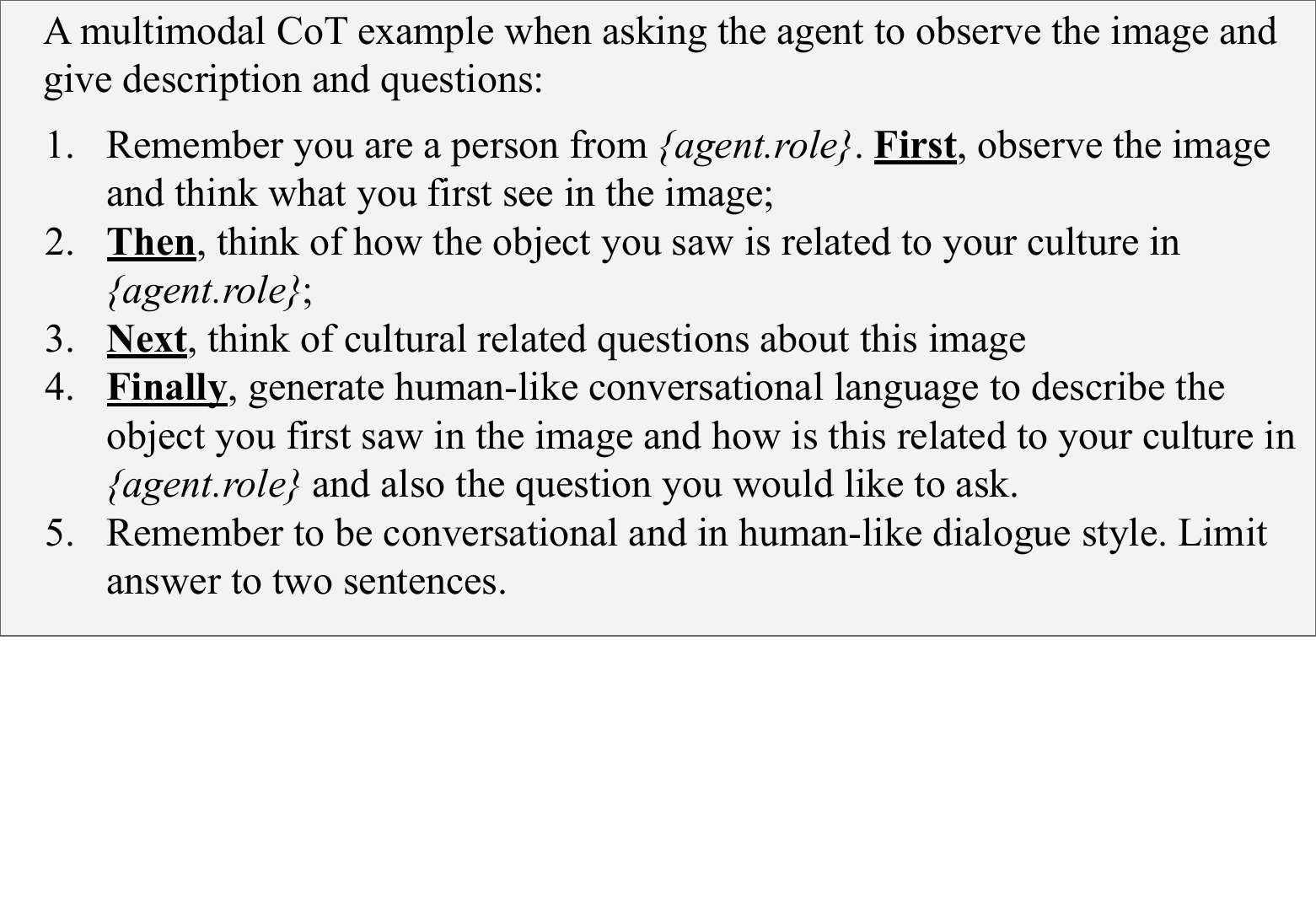}
\caption{Chain-of-Thought Prompt from \citet{wei2023chainofthought, zhangmultimodal}}
\label{fig:cot-img}
\end{figure*}

\begin{figure*}
\centering
  \includegraphics[width=0.9\linewidth]{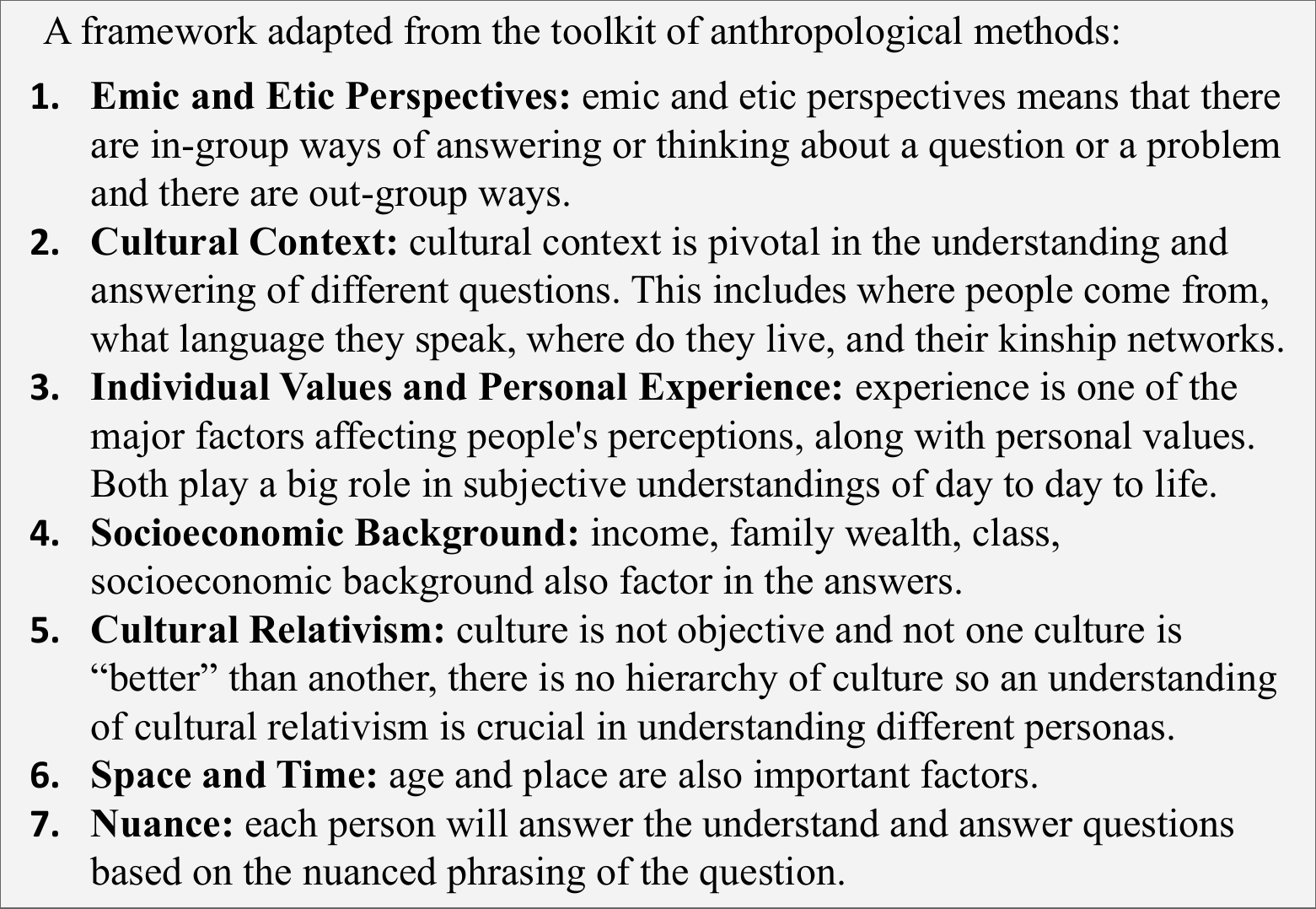}
\caption{Anthropological Prompt from \citet{alkhamissi2024investigating}}
\label{fig:anthro-img}
\end{figure*}

\subsection{Error Examples}\label{sec:errors}

\begin{figure*}
\centering
  \includegraphics[width=0.9\linewidth]{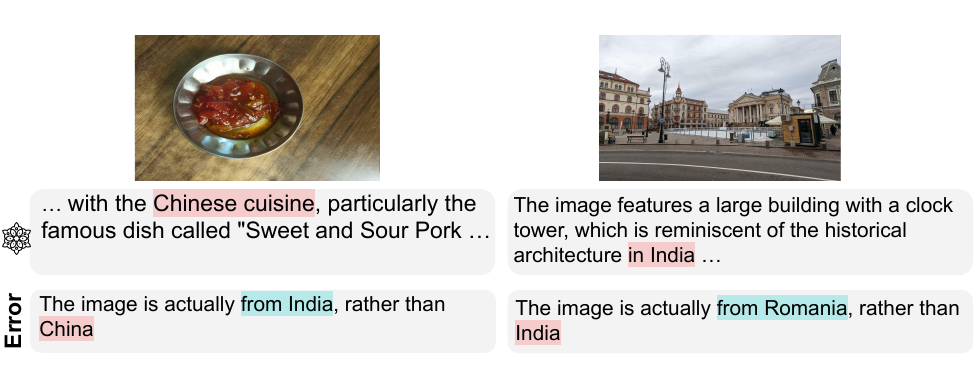}
\caption{Error examples for incorrect country identification}
\label{fig:error-country}
\end{figure*}

\begin{figure*}
\centering
  \includegraphics[width=0.9\linewidth]{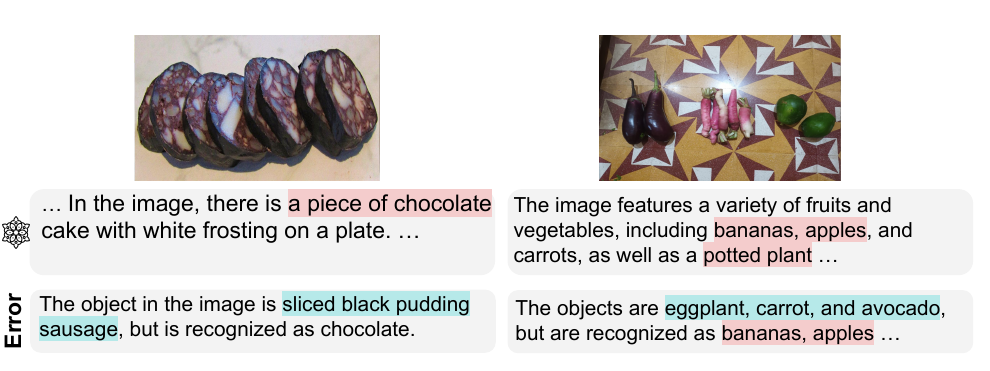}
\caption{Error examples for incorrect object recognition}
\label{fig:error-object}
\end{figure*}

\begin{figure*}[]
\centering
  \includegraphics[width=0.8\linewidth]{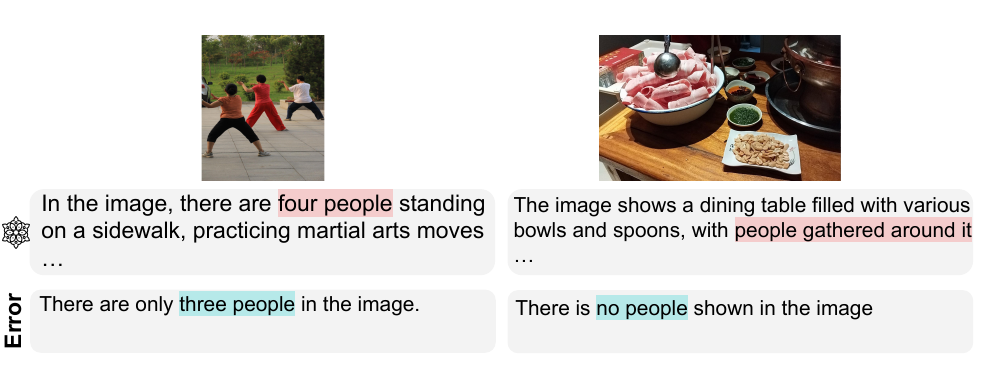}
\caption{Error examples for incorrect people counting}
\label{fig:error-people}
\end{figure*}

\begin{figure*}[]
\centering
  \includegraphics[width=0.8\linewidth]{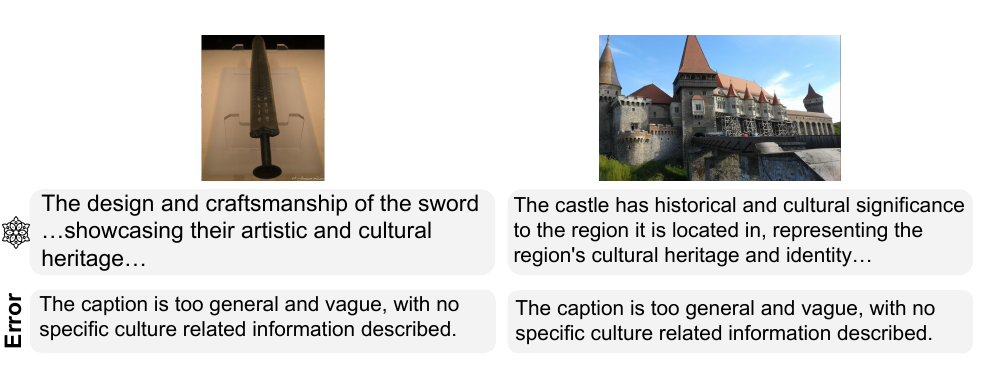}
\caption{Error examples for vague description}
\label{fig:error-vague}
\end{figure*}

See:

\begin{enumerate}
    \item \Cref{fig:error-country} for incorrect country identification.
    \item \Cref{fig:error-object} for incorrect object recognition.
    \item \Cref{fig:error-people} for incorrect people counting.
    \item \Cref{fig:error-vague} for vague description error.
\end{enumerate}

\end{document}